\documentclass[letterpaper]{article} 
\usepackage{aaai23}  
\usepackage{times}  
\usepackage{helvet}  
\usepackage{courier}  
\usepackage[hyphens]{url}  
\usepackage{graphicx} 
\usepackage{natbib}  
\usepackage{caption} 
\usepackage{algorithm2e}

\usepackage{selectp}
\usepackage{newfloat}
\usepackage{listings}
\DeclareCaptionStyle{ruled}{labelfont=normalfont,labelsep=colon,strut=off} 
\usepackage{amsmath}
\usepackage{multirow}
\usepackage{placeins}
\usepackage{subfigure}

\usepackage{enumitem}
\newtheorem{definition}{Definition}

\newcommand{\blue}[1]{\textcolor{blue}{#1}}

\usepackage{xcolor,colortbl}
\definecolor{LightCyan}{rgb}{0.88,1,1}
\definecolor{Gray}{gray}{0.85}

\newcolumntype{a}{>{\columncolor{Gray}}c}

\title{FairFed: Enabling Group Fairness in Federated Learning}

\author {
    Yahya H. Ezzeldin\equalcontrib,\textsuperscript{\rm 1}
    Shen Yan\equalcontrib, \textsuperscript{\rm 1}
    Chaoyang He, \textsuperscript{\rm 1}
    Emilio Ferrara, \textsuperscript{\rm 1}
    Salman Avestimehr \textsuperscript{\rm 1}
}
\affiliations {
    \textsuperscript{\rm 1} University of Southern California\\
    yessa@usc.edu, shenyan@usc.edu, chaoyang.he@usc.edu, emiliofe@usc.edu, avestime@usc.edu
}

\usepackage{bibentry}

\begin{document}

\maketitle

\begin{abstract}
Training ML models which are fair across different demographic groups is of critical importance due to the increased integration of ML in crucial decision-making scenarios such as healthcare and recruitment. 
Federated learning has been viewed as a promising solution for collaboratively training machine learning models among multiple parties while maintaining the privacy of their local data.
However, federated learning also poses new challenges in mitigating the potential bias against certain populations (e.g., demographic groups), as this typically requires centralized access to the sensitive information (e.g., race, gender) of each datapoint. 
Motivated by the importance and challenges of group fairness in federated learning, in this work, we propose FairFed, a novel algorithm for fairness-aware aggregation to enhance group fairness in federated learning. 
Our proposed approach is server-side and agnostic to the applied local debiasing thus allowing for flexible use of different local debiasing methods across clients. 
We evaluate FairFed empirically versus common baselines for fair ML and federated learning, and demonstrate that it provides fairer models particularly under highly heterogeneous data distributions across clients. 
We also demonstrate the benefits of FairFed in scenarios involving naturally distributed real-life data collected from different geographical locations or departments within an organization.
\end{abstract}

\section{Introduction}
An important notion of fairness in machine learning, \textit{group fairness}~\citep{dwork2012fairness}, concerns the mitigation of bias in the performance of a trained model against certain protected demographic groups, which are defined based on sensitive attributes within the population (e.g., gender, race). 
Several approaches to achieve group fairness have been studied in recent years in centralized settings. However, these approaches rely on the availability of the entire dataset at a central entity during training, and are therefore unsuitable for application in Federated Learning (FL).

Federated learning allows for decentralized training of large-scale models without requiring direct access to clients' data, hence maintaining their privacy \citep{Kairouz2021AdvancesAO,wang2021field}.
However, this decentralized nature makes it complicated to translate solutions for fair training from centralized settings to FL, where the decentralization of data is a major cornerstone. This gives rise to the key question that we attempt to answer in this paper: \textit{How can we train a classifier using FL so as to achieve group fairness, while maintaining data decentralization?}
\begin{figure}[t!]
\centering
    \includegraphics[width=0.85\columnwidth]{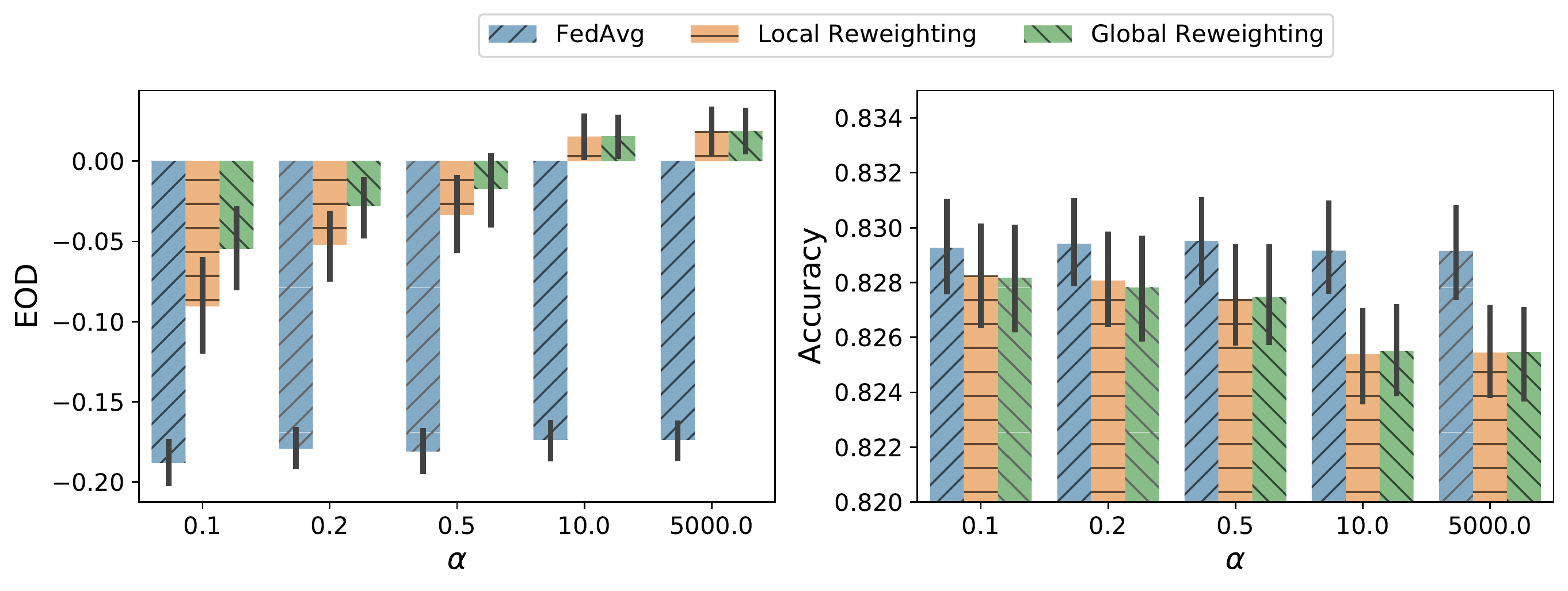}
    \vspace{-1em}
    \caption{Comparison of local/global debiasing under different heterogeneity levels $\alpha$. Smaller $\alpha$ indicates more heterogeneity across clients. For the Equal Opportunity Difference (EOD) metric, values closer to 0 indicate better fairness.}
    \vspace{-1.5em}
    \label{fig:local_compare}
\end{figure}
\subsubsection{Potential approaches for group fairness in FL.} 
One potential solution that one may consider for training fair models in FL is for each client to apply local debiasing on its locally trained models (without sharing any additional information or data), while the FL server simply aggregates the model parameters in each round using FL aggregation algorithms such as FedAvg~\citep{mcmahan2017communication} (or its subsequent derivatives, e.g., FedOPT \citep{reddi2020adaptive}, FedNova \citep{wang2020tackling}). 
Although this allows for training a global model without explicitly sharing the local datasets,
but the drawback is that applying a debiasing mechanism at each client in isolation on its local dataset can results in poor performance in scenarios where data distributions are highly-heterogeneous across clients (See Figure~\ref{fig:local_compare}). 

Another potential solution for fair training in FL would be to adapt a debiasing technique from the rich literature of centralized fair training to be used in FL. Although, this may result in reasonable fair training (see Figure~\ref{fig:local_compare}), but in the process of applying this debiasing globally, the clients may need to exchange additional detailed information with the server about their dataset constitution which can leak information about different subgroups in the client's dataset. For example, the server may require knowledge of the model's performance on each group in a client's dataset and/or local statistical information about each group in the dataset.

\subsubsection{The Proposed FairFed Approach.} 
\begin{figure}
    \centering
    \includegraphics[width=0.47\textwidth]{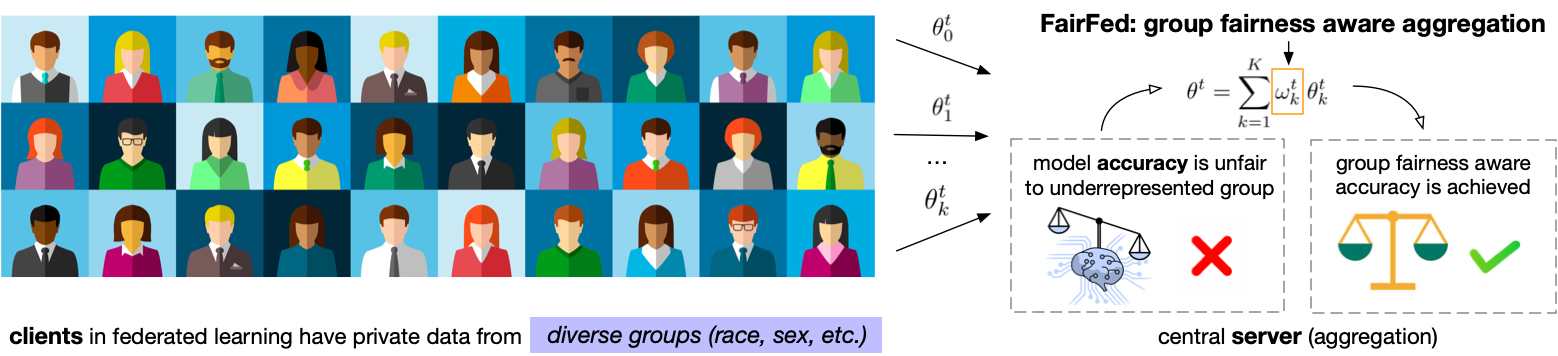}
    \vspace{-1.7em}
    \caption{FairFed: Group fairness-aware FL  framework.}
    \vspace{-1em}
\label{fig:overview}
\end{figure}

Motivated by the drawbacks of the two aforementioned directions, in this work, we propose \textit{FairFed}, a strategy to train fair models via a fairness-aware aggregation method (Figure~\ref{fig:overview}). In FairFed, each client performs \textit{local debiasing} on its own local dataset, thus maintaining data decentralization and avoiding exchange of any explicit information of its local data composition. To amplify the local debiasing performance, the clients will evaluate the \textit{fairness of the global model on their local datasets} in each FL round and collectively collaborate with the server to adjust its model aggregation weights. The weights are a function of the mismatch between the global fairness measurement (on the full dataset) and the local fairness measurement at each client, favoring clients whose local measures match the global fairness measure. We carefully design the exchange between the server and clients during weights computation, making use of the secure aggregation protocol (SecAgg)~\citep{bonawitz2017practical} to prevent the server from learning any explicit information about any single client's dataset. We present the details of FairFed in Section~\ref{sec:fairfed}.

The server-side/local debiasing nature of FairFed gives it the following benefits over existing fair FL strategies:
\begin{itemize}
    \item 
    \textbf{Enhancing group fairness under data heterogeneity:} One of the biggest challenges to group fairness in FL is the heterogeneity of data distributions across clients, which limits the impact of local debiasing efforts on the global data distribution. FairFed shows significant improvement in fairness performance under highly heterogeneous distribution settings and outperforms state-of-the-art methods for fairness in FL, indicating promising implications from applying it to real-life applications.
\item \textbf{Freedom for different debiasing across clients:} As FairFed works at the server side and only requires evaluation metrics of the model fairness from the clients, it is more flexible to run on top of heterogeneous client debiasing strategies (we expand on this notion in Sections~\ref{sec:fairfed}). For example, different clients can adopt different local debiasing methods based on the properties (or limitations) of their devices and data partitions.
\end{itemize}

\section{Background and Related Work}
\subsubsection{Group fairness in centralized learning.} In classical centralized ML, common approaches for realizing group fairness can be classified into three categories: pre-processing \citep{grgic2018beyond,feldman2015certifying}, in-processing \citep{kamishima2012fairness,zhang2018mitigating, roh2021fairbatch} and post-processing \citep{lohia2019bias, kim2019multiaccuracy} techniques. However, a majority of these techniques need centralized access to the sensitive information (e.g., race) of each datapoint, making it unsuitable for FL. As a result developing effective approaches for fair FL is an important area of study.

\subsubsection{Fairness in federated learning.}
New challenges in FL have introduced different notions of fairness. These new notions include for example, \textit{client-based fairness} \citep{li2019fair,mohri2019agnostic} which aims to equalize model performance across different clients or \textit{collaborative fairness} \citep{lyu2020collaborative,Wang2021FederatedLW} which aims to reward a highly-contributing participant with a better performing local model than is given to a low-contributing participant. In this paper, we instead focus on the notion of \textit{group fairness} in FL, where each datapoint in the FL system belongs to particular group, and we aim to train models that do not discriminate against any group of datapoints. 

Several recent works have made progress on group fairness in  FL. One common research direction is to distributively solve an optimization objective with fairness constraints \citep{zhang2020fairfl,du2021fairness,galvez2021enforcing}, which requires each client to share the statistics of the sensitive attributes of its local dataset to the server. The authors in~\citep{abay2020mitigating} investigated the effectiveness of adopting a global reweighting mechanism. In~\citep{anonymous2022improving}, an adaptation of the FairBatch debiasing algorithm~\citep{roh2021fairbatch} is proposed for FL where clients use FairBatch locally and the weights are updated through the server in each round. 
In~\citep{papadaki2021federating}, an algorithm is proposed to achieve minimax fairness in federated learning. 
In these works, the server requires each client to explicitly share the performance of the model on each subgroup separately; for example (males with +ve outcomes, females with +ve outcomes, etc).
Differently from these works, our proposed FairFed method does not restrict the local debiasing strategy of the participating clients, thus increasing the flexibility of the system. Furthermore, FairFed does not share explicit information on the model performance for any specific group within a client's dataset. Finally, our empirical evaluations consider extreme cases of data heterogeneity and demonstrate that our method can yield significant fairness improvements in these situations.

\section{Preliminaries}\label{sec:prelim}
We begin by reviewing the standard FL setup~\citep{mcmahan2017communication}, and then introduce key definitions and metrics for group fairness. We then extend these to the FL setting by defining the notions of global and local fairness in FL.
\subsection{Federated Learning Setup}
Following a standard FL setting~\cite{mcmahan2017communication}, we consider a scenario where $K$ clients collaborate with a server to find a parameter vector $\theta$ that minimizes the weighted average of the loss across all clients. In particular:
\vspace{-1em}
\begin{equation}
\min_{\theta}f(\theta) = \sum_{k=1}^K\omega_k L_k(\theta),
\label{eq:objective_fl}
\end{equation}
where: $L_k(\theta)$ denotes the local objective at client $k$; $\omega_k \geq 0$, and $\sum{\omega_k} = 1$. The local objective $L_k$’s can be
defined by empirical risks over the local dataset $\mathcal{D}_k$ of size $n_k$ at client $k$, i.e., $L_k(\theta) = \frac{1}{n_k}\sum_{(\mathbf{x},y) \in \mathcal{D}_k}\ell(\theta,\mathbf{x},y)$.

To minimize the objective in~\eqref{eq:objective_fl}, the federated averaging algorithm FedAvg, proposed in~\citep{mcmahan2017communication}, samples a subset of the $K$ clients per round to perform local training of the global model on their local datasets. The model updates are then averaged at the server, being weighted based on the size of their respective datasets. 

To ensure the server does not learn any information about the values of the individual transmitted updates from the clients beyond the aggregated value that it sets out to compute, FedAvg typically employs a Secure Aggregation (SecAgg) algorithm ~\cite{bonawitz2017practical}.

Training using FedAvg and its subsequent improvements (e.g., FedOPT \citep{reddi2020adaptive}) allows training of a high-performance global model, however, this collaborative training can result in a global model that discriminates against an underlying demographic group of datapoints (similar to biases incurred in centralized training of machine learning models~\citep{dwork2012fairness}). We highlight key notions of group fairness in fair ML in the following subsection.

\subsection{Notions of group fairness}

In sensitive machine learning applications, a data sample often contains private and sensitive demographic information that can lead to discrimination.
In particular, we assume that each datapoint is associated with a sensitive binary attribute $A$ (e.g., gender or race). For a binary prediction model $\hat{Y}(\theta, \mathbf{x})$, the fairness is evaluated with respect to its performance compared to the underlying groups defined by the sensitive attribute $A$. We use $A=1$ to represent the privileged group (e.g., male), while $A=0$ is used to represent the under-privileged group (e.g., female). For the binary model output $\hat{Y}$ (and similarly the label $Y$), $\hat{Y}=1$ is assumed to be the positive outcome. 
Using these definitions, we can now define two group fairness notions that are applied in group fairness literature for centralized training:
\begin{definition}[Equal Opportunity] :
Equal opportunity \citep{hardt2016equality} measures the performance a binary predictor $\hat{Y}$ with respect to $A$ and $Y$. The predictor is considered fair from the equal opportunity perspective if the true positive rate is independent of the sensitive attribute $A$.
To measure this, we use the Equal Opportunity Difference (EOD), defined as
\begin{align}
 &EOD {=} \Pr(\hat{Y}{=}1{|}A{=}0{,}Y{=}1\!) {-} \Pr(\hat{Y}{=}1\!{|}A{=}1{,}Y{=}1\!).\!
 \label{eq:EOD}
\end{align}
\end{definition}

\begin{definition}[Statistical Parity] :
Statistical parity \citep{dwork2012fairness} rewards the classifier for classifying each group as positive at the same rate. Thus, a binary predictor $\hat{Y}$ is fair from the statistical parity perspective if    
$\Pr(\hat{Y}=1|A=1)=\Pr(\hat{Y}=1|A=0)$. Thus, the Statistical Parity Difference (SPD) metric is defined as 
\begin{equation}
SPD = \Pr(\hat{Y}=1|A=0) - \Pr(\hat{Y}=1|A=1).
\label{eq:SPD}
\end{equation}
\end{definition}
For the EOD and SPD metrics, values closer to zero indicate better fairness. Positive fairness metrics indicate that the unprivileged group outperform the privileged group.

\subsection{Global vs local group fairness in FL}
The fairness definitions above can be readily applied to centralized model training to evaluate the performance of the trained model. However, in FL, clients typically have non-IID data distributions, which gives rise to different levels of consideration for fairness: \emph{global fairness} and \emph{local fairness}. 

The {\bf global fairness} of a given model considers the full dataset $\bar{\mathcal{D}} = \cup_k \mathcal{D}_k$ across the $K$ clients in FL, which is our end-goal in fair FL; to train a model that is in general non-discriminatory to any group in the global dataset.  

In contrast, when only the local dataset $\mathcal{D}_k$ at client $k$ is considered, we can define the {\bf local fairness} performance by applying~\eqref{eq:EOD} or~\eqref{eq:SPD} on the data distribution at client $k$.

We can highlight the difference between global and local fairness using the example of the \textit{EOD} metric. 
For a classifier $\hat{Y}$, the global fairness EOD metric $F_{global}$ is given by 
\begin{align}
     \!\!F_{global} {=} \Pr(\hat{Y}{=}1{|}A{=}0{,}Y{=}1\!) {-} \Pr(\hat{Y}{=}1{|}A{=}1{,}Y{=}1\!),\!
     \label{eq:global_metric_eq}
\end{align}
where the probability above is based on the full dataset distribution (a mixture of the distributions across the clients). 
In contrast, the local fairness metric $F_k$ at client $k$ is 
\begin{align}
     F_k = &\Pr(\hat{Y}\!=\!1|A\!=\!g,Y\!=\!1,C\!=\!k) \nonumber \\
     & - \Pr(\hat{Y}\!=\!1|A\!=\!g,Y\!=\!1,C\!=\!k),
     \label{eq:local_metric_eq}
\end{align}
where the parameter $C=k$ denotes that the $k$-th client (and dataset $\mathcal{D}_k$) is considered in the fairness evaluation.

Note that if clients have IID distributions (i.e., distributions that are independent of $C$), global and local fairness match. However, they can greatly differ in the non-IID case.

\RestyleAlgo{ruled}
\SetAlgoNoEnd
\begin{algorithm*}[!t]\small
\caption{FairFed Algorithm (tracking EOD)
}\label{alg:fairagg}
 \underline{Initialize}: global model parameter $\theta_0$ and weights $\{\omega_k^0\}$ as $\omega_k^0 = n_k/\sum_{i=1}^K n_i$, $\forall k \in [K]$\;
\underline{Dataset statistics}:
Aggregate statistics $\mathcal{S} \!=\! \left\{\ \Pr(A{=}1,Y{=}1), \Pr(A{=}0,Y{=}1)\right\}$ from clients using Secure Aggregation (SecAgg) and send it to clients\;
\For {each round $t=1,2,\cdots $ }
{   
      \!$F_{global}^t\ ,\ \overline{Acc^t} \gets \text{SecAgg}\left(\left\{\text{\bf ClientLocalMetrics}\left(k,\theta^{t-1}\right)\right\}_{k=1}^K\right)$; \blue{\emph{// SecAgg to get $\overline{Acc^t}$ and global fairness $F_{global}^t$ as in~\eqref{eq:global_component}}}\;
    {$\frac{1}{K} \sum_i \Delta_i \!\gets \text{SecAgg}\left(\left\{\text{\bf  ClientMetricGap}\left(k,\theta^{t-1},F_{global}^t,\overline{Acc^t}\right)\right\}_{k=1}^K\right)$}; \emph{\blue{// SecAgg to compute mean of metric gaps}}\;
    \blue{\emph{// Compute aggregation weights locally at clients based on~\eqref{eq:FairAgg_weight_def} then use SecAgg to aggregate weighted local model updates}}\;
    {
    $\left(\sum_{k=1}^K\bar{\omega}_k^t\theta_k^{t}\right)\ ,\  \left(\sum_{k=1}^K\bar{\omega}_k^t\right) \gets \text{SecAgg}\left(\left\{\text{\bf ClientWeightedModelUpdate}\left(k,\theta^{t-1},\omega_k^t, \frac{1}{K} \sum_i \Delta_i \right)\right\}_{k=1}^K\right)$\;
    $\theta^{t+1} \gets \left(\sum_{k=1}^K\bar{\omega}_k^t\theta_k^{t}\right) / \left(\  \sum_{k=1}^K\bar{\omega}_k^t\right)$}\;
}
~
\end{algorithm*}

\section{FairFed: Fairness-aware aggregation in FL}\label{sec:fairfed}
In this section, we introduce our proposed approach FairFed which uses local debiasing due for its advantages for data decentralization, while addressing its challenges by adjusting how the server aggregates local model updates from clients.

\subsection{Our proposed approach (FairFed)} 
Recall that in the $t$-th round in FedAvg~\citep{mcmahan2017communication}, local model updates $\{\theta_k^t\}_{k=1}^K$ are weight-averaged to get the new global model parameter $\theta^t$ as:
$\theta^{t+1} = \sum_{k=1}^K \omega_k^t \ \theta_k^t$,
where the weights $\omega_k^t = n_k / \sum_k n_k$ depend only on the number of datapoints at each client.

As a result, a fairness-oblivious aggregation would favor clients with more datapoints. If the training at these clients results in locally biased models, then the global model can potentially be biased since the weighted averaging exaggerates the contribution of model update from these clients.

Based on this observation, in FairFed, we propose a method to optimize global group fairness $F_{global}$ via adaptively adjusting the aggregation weights of different clients based on their local fairness metric $F_k$.
In particular, given the current global fairness metric $F_{global}^t$ (we will discuss later in the section, how the server can compute this value), then in the next round, the server gives a slightly higher weight to clients that have a similar local fairness $F_k^t$ to the global fairness metric, thus relying on their local debiasing to steer the next model update towards a fair global model.

Next, we detail how FairFed computes the aggregation weights in each round. 
The steps performed while tracking the EOD metric in FairFed are shown in Algorithm~\ref{alg:fairagg}.

\subsection{Computing aggregation weights for FairFed}
At the beginning of training, we start with the default FedAvg weights $\omega_k^0 = n_k /\sum_{k=1}^K n_k$. Next, in each round $t$, we update the weight assigned to the $k$-th client based on the current gap between its local fairness metric $F_k^t$ and the global fairness metric $F_{global}$. In particular, the weight update follows this formula $\forall k \in [K]$:
\begin{align}\label{eq:FairAgg_weight_def}
    \Delta_k^t &=
    \begin{cases}
      \left|Acc_k^t - \overline{Acc^t}\right| & \text{if $F_k^t$ is undefined,} \\ 
      |F^t_{\rm global} - F_k^t| & \text{otherwise}
    \end{cases}, \nonumber
    \\
    \bar{\omega}_k^t &{=} \bar{\omega}_k^{t-1} {-} \beta \!\!\left(\!\!\Delta_k - \frac{1}{K}\!\sum_{i=1}^K \Delta_i\!\!\right)\!,\quad  {\omega}_k^t = \frac{\bar{\omega}_k^t}{ \textstyle\sum_{i=1}^K \bar{\omega}_i^t}. 
\end{align}
where: (i) $Acc_k^t$ represents the local accuracy at client $k$, and $\overline{Acc^t} = \sum_{k=1}^K Acc_k\times n_k/\sum_{k=1}^K n_k$, is global accuracy across the full dataset, respectively; (ii) $\beta$ is a parameter that controls the fairness budget for each update, thus impacting the trade-off between model accuracy and fairness. Higher values of $\beta$ result in fairness metrics having a higher impact on the model optimization, while a lower $\beta$ results in a reduced perturbation to the default FedAvg weights due to fair training; note that at $\beta=0$, FairFed is equivalent to FedAvg, as the initial weights $\omega_k^0$ are unchanged.

\subsubsection{Intuition for weight update.} The intuition behind the update in~\eqref{eq:FairAgg_weight_def} is to effectively rank the clients in the FL system based on how far their local view of fairness (measured through their local metric) compares to the global fairness metric; closer views to the global metric get assigned higher weights while clients with local metrics that significantly deviate from the global metric will have their weights reduced. The significance is decided based on whether the gap from the global metric is above the average gap (across clients) or vice versa.
Note that, whenever, the client distribution makes the local metric $F_k$ undefined\footnote{For instance, in the case of the EOD metric, this happens whenever $\Pr(A=1,Y=1) = 0$ or $\Pr(A=0,Y=1)=0$.}, FairFed relies on the discrepancy between the local and global accuracy metric as a proxy to compute the fairness metric gap $\Delta_k$.

Thus, so far, the training process of FairFed at each iteration follows the following conceptual steps: 
\begin{enumerate}
    \item Each client computes its updated local model parameters;
    \item The server computes the global fairness metric value $F_{global}^t$ and global accuracy $\overline{Acc^t}$ using secure aggregation and broadcasts them back to the clients; 
    \item Each client computes its metric gap $\Delta_k^t$ and from it, it calculates its aggregation weight ${\omega}_k^t$ with the help of the server as defined in~\eqref{eq:FairAgg_weight_def};
    \item The server next aggregates the weighted local updates $\omega_k^t \theta_k^t$ using secure aggregation to compute the new global model and broadcasts it to the clients.
\end{enumerate}
A detailed description on performing these steps using secure aggregation (SecAgg) is shown in Algorithm~\ref{alg:fairagg}, where SecAgg($\{b_i\}_{i=1}^K$) computes $\sum_i b_i$ using secure aggregation.

\subsubsection{Flexibility of FairFed with heterogeneous debiasing.} Note that the FairFed weights $\omega_k^t$ in~\eqref{eq:FairAgg_weight_def} rely only on the global and local fairness metrics and are not tuned towards a specific local debiasing method. Thus, FairFed is flexible to be applied with different debiasing methods at each client, and the server will incorporate the effects of these different methods by reweighting their respective clients based on their local/global metrics and  the weight computation in~\eqref{eq:FairAgg_weight_def}.

\subsection{How the server gets the global metric $F_{global}$}\label{sec:computing_global_metric_EOD}
One central ingredient for computing weights in FairFed is the server's ability to calculate the global metric $F_{global}$ in each round (recall equation~\eqref{eq:FairAgg_weight_def}) without the clients having to share their datasets with the server or any explicit information about their local group distribution. If the metric of interest is EOD, we next show how the server can compute $F_{global}$ from the clients using secure aggregation. Similar computations follow for SPD and are presented in Appendix~\ref{app:SPD}.
Let $n = \sum_{k=1}^K n_k$; the EOD metric in~\eqref{eq:global_metric_eq} can be rewritten as:
\begin{align}
\label{eq:global_component}
    &F_{global} =\Pr(\hat{Y}\!=\!1|A\!=\!0,Y\!=\!1) \!-\! \Pr(\hat{Y}\!=\!1|A\!=\!1,Y\!=\!1)\nonumber\\
 &\!\!=\!\!\sum_{k=1}^{K}\!\frac{n_k}{n} \Bigg[ \tfrac{\Pr(\hat{Y}=1|A=0,Y=1,C=k)\Pr(\!A=0,Y=1|C=k)}{\Pr(Y=1,A=0)} \nonumber \\
 &\hspace{2.5em}\underbrace{\hspace{1.5em} - \tfrac{\Pr(\hat{Y}=1|A=1,Y=1,C=k)\Pr(\!A=1,Y=1|C=k)}{\Pr(Y=1,A=1)}\Bigg]}_{m_{global,k}}\!,
\end{align}
where $m_{global,k}$ is the summation component that each client $k$ computes locally. Thus, the global EOD metric $F_{global}$ can be computed at the server by applying secure aggregation~\citep{bonawitz2017practical} to get the sum of the $m_{global,k}$ values from the $K$ clients without the server learning any information about the individual $m_{global,k}$ values.

Note that the conditional probabilities defining $m_{global,k}$ in~\eqref{eq:global_component} are all local performance metrics that can easily be computed locally by client $k$ using its local dataset $\mathcal{D}_k$. 
The only non-local terms in ${m_{global,k}}$ are the full dataset statistics $\mathcal{S} = \{\Pr(Y{=}1,A{=}0), \Pr(Y{=}1,A{=}1)\}$.
These statistics $\mathcal{S}$ can be aggregated at the server using a single round of secure aggregation (e.g.,~\citep{bonawitz2017practical}) at the start of training, then be shared with the clients to enable them to compute their global fairness component ${m_{global,k}}$. 

\section{Experimental Evaluation}\label{sec:experimental_evaluation}
In this section, we investigate the performance of FairFed under different system settings. In particular, we explore how the performance changes with different heterogeneity levels in data distributions across clients. We also evaluate how the trade-off between fairness and accuracy changes with the fairness budget $\beta$ in FairFed (see equation~\eqref{eq:FairAgg_weight_def}). Additional experiments in Appendix~\ref{app:more_results} investigate how the performance of FairFed changes when different local debiasing approaches are used across clients. 

\subsection{Experimental setup}
\label{sec:config}
\subsubsection{Implementation.} We developed FairFed using \texttt{FedML} \citep{he2020fedml}, which is a research-friendly FL library for exploring new algorithms. We use a server with \texttt{AMD EPYC 7502 32-Core} CPU Processor, and use a parallel training paradigm, where each client is handled by an independent process using MPI (message passing interface).

\subsubsection{Datasets.} 
In this section, we demonstrate the performance of different debiasing methods using two binary decision datasets that are widely investigated in fairness literature: the Adult~\citep{Dua:2019} dataset and ProPublica COMPAS dataset~\citep{larson2016we}. 
In the \textbf{Adult} dataset~\citep{Dua:2019}, we predict the yearly income (with binary label: over or under \$50,000) using twelve categorical or continuous features. The gender (defined as male or female) of each subject is considered as the sensitive attribute. 
For the ProPublica \textbf{COMPAS} dataset relates to recidivism, which is to assess if a criminal defendant will commit an offense within a certain future time. 
Features in this dataset include the number of prior offenses, age of the defendant, etc. The race (classified as white or non-white) of the defendant is the sensitive attribute of interest.

\subsubsection{Configurable data heterogeneity for diverse sensitive attribute distributions.}
To understand the performance of our method and the baselines, under different distributions of the sensitive attribute across clients, a configurable data synthesis method is needed. In our context, we use a generic non-IID synthesis method based on the Dirichlet distribution proposed in~\citep{sampling_non_iid_google} but apply it in a novel way for \textit{configurable sensitive attribute distribution}: for each sensitive attribute value $a$, we sample $\mathbf{p}_a\sim{\rm Dir}(\alpha)$ and allocate a portion $p_{a,k}$ of the datapoints with $A=a$ to client $k$.
The heterogeneity of the distributions across clients is controlled via $\alpha$, where $\alpha \to \infty$ results in IID distributions. Examples of these heterogeneous distributions for the Adult and COMPAS datasets are shown in Appendix~\ref{app:datasets}.

\smallskip

\begin{table*} \small
\centering
    \begin{tabular}{c|l |a c c c c | a c c c c}
    \hline
    \hline
       & \multirow{3}{*}{Method}& \multicolumn{5}{c|}{Adult ($\beta = 1$)} & \multicolumn{5}{c}{COMPAS ($\beta = 1$)}  \\
       \cline{3-12}
          &    & \multicolumn{5}{c|}{Heterogeneity Level $\alpha$} & \multicolumn{5}{c}{Heterogeneity Level $\alpha$}\\
         & & 0.1 & 0.2 & 0.5 & 10 & 5000 & 0.1 & 0.2 & 0.5 & 10 & 5000 \\
         \hline
           \multirow{7}{*}{Acc.} 
            & FedAvg &  \textbf{0.835} &  \textbf{0.836} &  \textbf{0.835} &  \textbf{0.836} &  \textbf{0.837} &  \textbf{0.674} &  \textbf{0.673} &  \textbf{0.675} &  \textbf{0.674} &  \textbf{0.675} \\
            & Local / [Best] &  0.831 &  0.833 & 0.834 &  0.831 & 0.829 & 0.666 &  0.659 &  0.665 &  0.663 &  0.664 \\
            & Global RW &  0.834 &  0.833 &  0.831 &  0.829 &  0.829 &  0.673 &  0.671 &  0.672 &  0.676 &  0.675 \\ 
            & FedFB &  0.825 &  0.825 &  0.829 &  0.832 &  0.832 &  0.674 &  0.673 &  0.675 &  0.677 &  0.677 \\
            & FairFed / RW &  0.830 &  0.834 &  0.832 &  0.829 &  0.829 &  0.672 &  0.670 &  0.669 &  0.669 &  0.673 \\
            & FairFed / FairRep &  0.824 &  0.833 &  0.834 &  0.834 &  0.834 &  0.661 &  0.655 &  0.663 &  0.663 &  0.660 \\
            & FairFed / FairBatch &  0.829 &  0.833 &  0.830 &  0.830 &  0.831 &  0.659 &  0.664 &  0.665 &  0.661 &  0.661 \\         
        \hline
        \multirow{7}{*}{EOD} 
        & FedAvg & -0.174 & -0.173 & -0.176 & -0.179 & -0.180 & -0.065 & -0.071 & -0.067 & -0.076 & -0.078 \\
        & Local / [Best] &  0.052 & -0.009 & -0.006 & -0.013 & 0.014 & -0.055 & -0.051 & -0.054 & -0.038 & -0.035\\
        & Global RW & -0.030 &  0.019 &  0.022 &  0.017 &  0.010 & -0.060 & -0.065 & -0.066 & -0.076 & -0.077 \\
        & FedFB & -0.019 &  0.015 &  0.015 & -0.012 & -0.012 & -0.062 & -0.061 & -0.063 & -0.077 & -0.072 \\ 
        \rowcolor{LightCyan}& FairFed / RW & \textbf{-0.017} &  \textbf{0.001} &  0.018 &  0.016 &  0.013 & -0.057 & -0.065 & -0.053 & -0.067 & -0.061 \\
        \rowcolor{LightCyan}& FairFed / FairRep &  0.023 & -0.009 & -0.071 & -0.174 & -0.187 &  \textbf{0.037} &  \textbf{0.023} &  \textbf{0.043} &  0.046 &  0.039 \\
        \rowcolor{LightCyan}& FairFed / FairBatch & -0.020 &  \textbf{0.001} &  \textbf{0.000} & \textbf{-0.005} & \textbf{-0.004} & -0.048 & -0.048 & -0.049 & \textbf{-0.035} & \textbf{-0.031} \\        
          \hline
          \multirow{7}{*}{SPD} 
            & FedAvg & -0.176 & -0.179 & -0.178 & -0.177 & -0.177 & -0.137 & -0.141 & -0.138 & -0.142 & -0.149 \\
            & Local / [Best] & -0.107 & -0.128 & -0.122 & -0.122 & -0.108 & -0.124 & -0.114 & -0.121 & -0.101 & -0.103\\
            & Global RW & -0.123 & -0.111 & -0.106 & \textbf{-0.107} & \textbf{-0.108} & -0.131 & -0.137 & -0.135 & -0.140 & -0.147 \\
            & FedFB & \textbf{-0.093} & \textbf{-0.100} & \textbf{-0.103} & -0.157 & -0.156 & -0.133 & -0.131 & -0.135 & -0.147 & -0.144 \\
           & FairFed / RW & -0.116 & -0.119 & -0.111 & \textbf{-0.107} & \textbf{-0.108} & -0.129 & -0.135 & -0.127 & -0.132 & -0.135 \\
           & FairFed / FairRep & -0.123 & -0.128 & -0.148 & -0.181 & -0.185 & \textbf{-0.018} &\textbf{ -0.025} & \textbf{-0.011} & -0.019 & \textbf{-0.019} \\
           & FairFed / FairBatch & -0.116 & -0.117 & -0.113 & -0.117 & -0.119 & -0.112 & -0.114 & -0.116 & \textbf{-0.016} & -0.026 \\
    \hline
    \hline
    \end{tabular}    
    \caption{Performance comparison of data partition with different heterogeneity levels $\alpha$. A smaller $\alpha$ indicates a more heterogeneous distribution across clients. We report the average of 20 random seeds. For EOD and SPD metrics, values closer to zero indicate better fairness. Positive fairness metrics indicate that the unprivileged group outperform the privileged group. For brevity, we report the values achieved by the best local debiasing baseline (without FairFed) as Local / [Best] in the table.
    }
    \label{tab:compare_partition}
\end{table*}
\subsubsection{Baselines.}
We adopt the following state-of-the-art solutions as baselines:
\begin{itemize}[leftmargin=12pt]
\item 
\textbf{FedAvg }\citep{mcmahan2017communication}\textbf{:} the original FL algorithm for distributed training of private data. It does not consider fairness for different demographic groups.
\item 
\textbf{FedAvg + Local reweighting [Local / RW]:} Each client adopts the reweighting strategy \citep{kamiran2012data} to debias its local training data, then trains local models based on the pre-processed data. FedAvg is used to aggregate the local model updates at the server.
\item 
\textbf{FedAvg + FairBatch [Local / FairBatch]:} Each client adopts the state-of-the-art FairBatch in-processing debiasing strategy \citep{roh2021fairbatch} to debias its local training data and then aggregation uses FedAvg.
\item 
\textbf{FedAvg + Fair Linear Representation [Local / FairRep]:} Each client adopts the Fair Linear Representations pre-processing debiasing strategy \citep{he2020geometric} locally and aggregates using FedAvg.
\item 
\textbf{FedAvg + Global reweighting [Global RW]} \citep{abay2020mitigating}\textbf{:} A differential-privacy approach to collect noisy statistics such as the number of samples with privileged attribute values ($A{=}1$) and favorable labels ($Y{=}1$) from clients. Server computes global weights based on the collected statistics and shares them with the clients, which assign them to their data samples during FL training~\footnote{We apply the global reweighting approach in~\cite{abay2020mitigating} without differential-privacy noise in order to compare with the optimal debiasing performance of global reweighting.}.
\item 
\textbf{FedFB} \citep{anonymous2022improving}\textbf{:} An in-processing debiasing approach in FL based on FairBatch~\cite{roh2021fairbatch}. The server computes new weights for each group based on information from the clients in each round and broadcast them back to the clients. For fair comparison, we use FedFB that is optimized w.r.t EOD as in FairFed.

\end{itemize}

\subsubsection{Hyperparameters.} In this section, we use our FairFed approach and the above baselines to train a logistic regression model for binary classification tasks on the Adult and COMPAS datasets. All results are selected from the best EOD metric (closest to zero) obtained from grid search on the  important hyperparameters after averaging over 20 random seeds. In particular, for all baselines and FairFed, we consider learning rates in the set \{1e-3, 1e-2, 1e-1\}. For FairBatch based baselines~\cite{roh2021fairbatch} and FedFB~\cite{anonymous2022improving}, we consider learning rates for weight update in \{1e-3, 5e-3, 1e-2, 2e-2\}. Unless otherwise specified, we use a fairness budget of $\beta =1$ in FairFed. 

\subsection{Experimental results}
\label{sec:experiment}

\subsubsection{Performance under heterogeneous sensitive attribute distributions.}
We compared the performance of FairFed when used with three local debiasing methods (reweighting, FairRep~\cite{he2020geometric} and FairBatch~\cite{roh2021fairbatch}) against the baselines described in Section~\ref{sec:config}, under different heterogeneity levels.
The results are summarized in Table \ref{tab:compare_partition}. 

FairFed outperforms the baselines at different heterogeneity levels, but at highly homogeneous data distributions (i.e., a large $\alpha$ value), FairFed does not provide significant gains in fairness performance compared to local debiasing methods (except when using FairBatch). This is due to the fact that under homogeneous sampling, the distributions of the local datasets are statistically similar (and reflect the original distribution with enough samples), resulting in similar debiasing effect being applied across all clients when using the pre-processing methods (reweighting and FairRep). 
For a higher level of heterogeneity (i.e., at lower $\alpha = 0.1$), FairFed can improve EOD in Adult and COMPAS data by 93\% and 50\%, respectively. This is done at the expense of only a 0.3\% decrease in accuracy for both Adult and COMPAS datasets.
In contrast, at the same heterogeneity level, local strategies can only improve EOD by 65\% and 15\% for Adult and COMPAS datasets, respectively. Global reweighting  only improves EOD by 73\% and 2\% for Adult and COMPAS datasets, respectively. On average, across different heterogeneity levels, FedFB gives the performance closest to FairFed. For $\alpha=0.1$, FedFB improves EOD by 87\% and  1.5\% for Adult and COMPAS, respectively. 

Note, however, that FedFB requires the clients to share explicit information about the performance of the the model on each local subgroup in order to update the weights in FariBatch~\cite{anonymous2022improving}, which can potentially leak information about the clients' local datasets.

\begin{figure}[!t]
\centering
    \subfigure{
    \includegraphics[width=0.47\columnwidth]{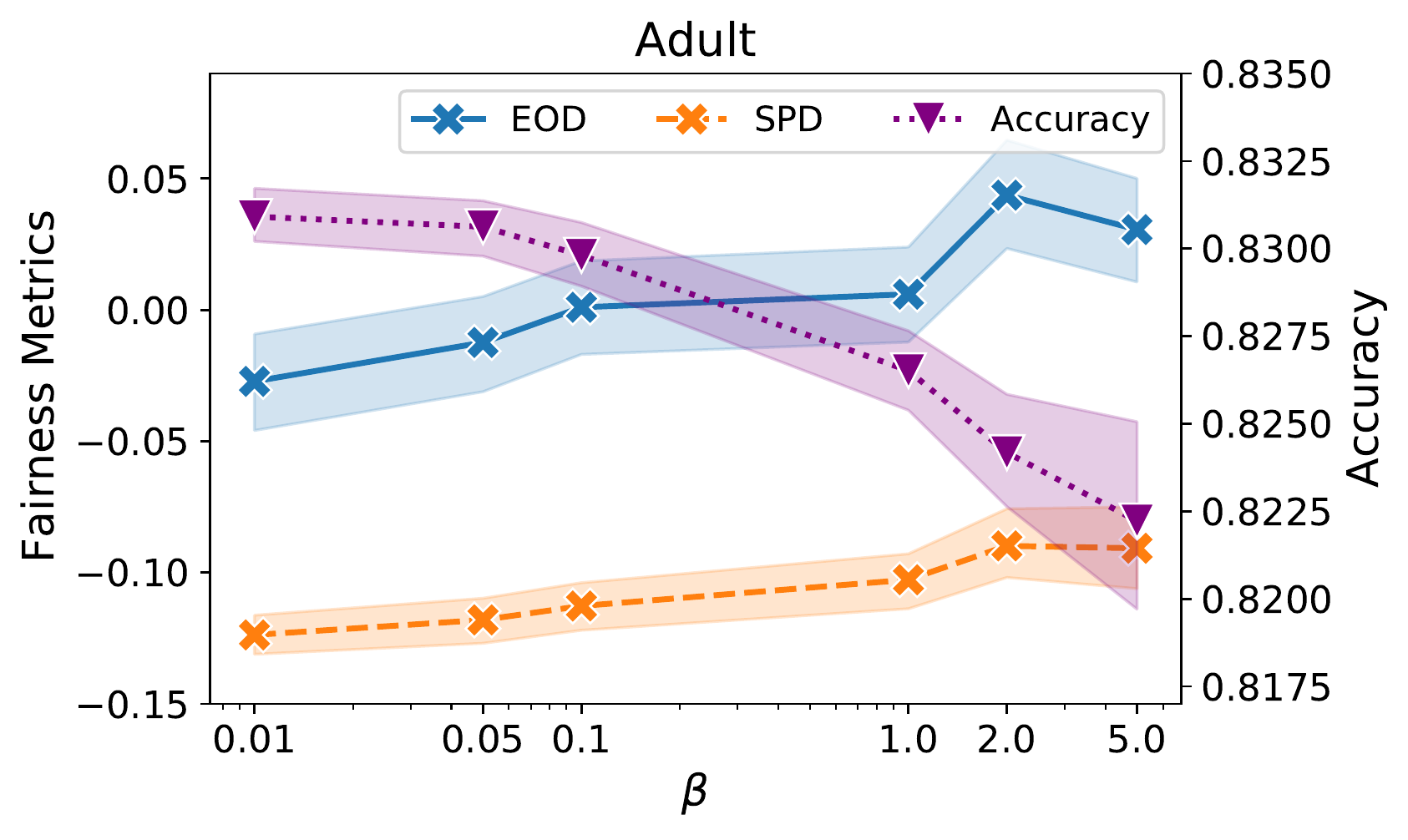}}
    \subfigure{
    \includegraphics[width=0.47\columnwidth]{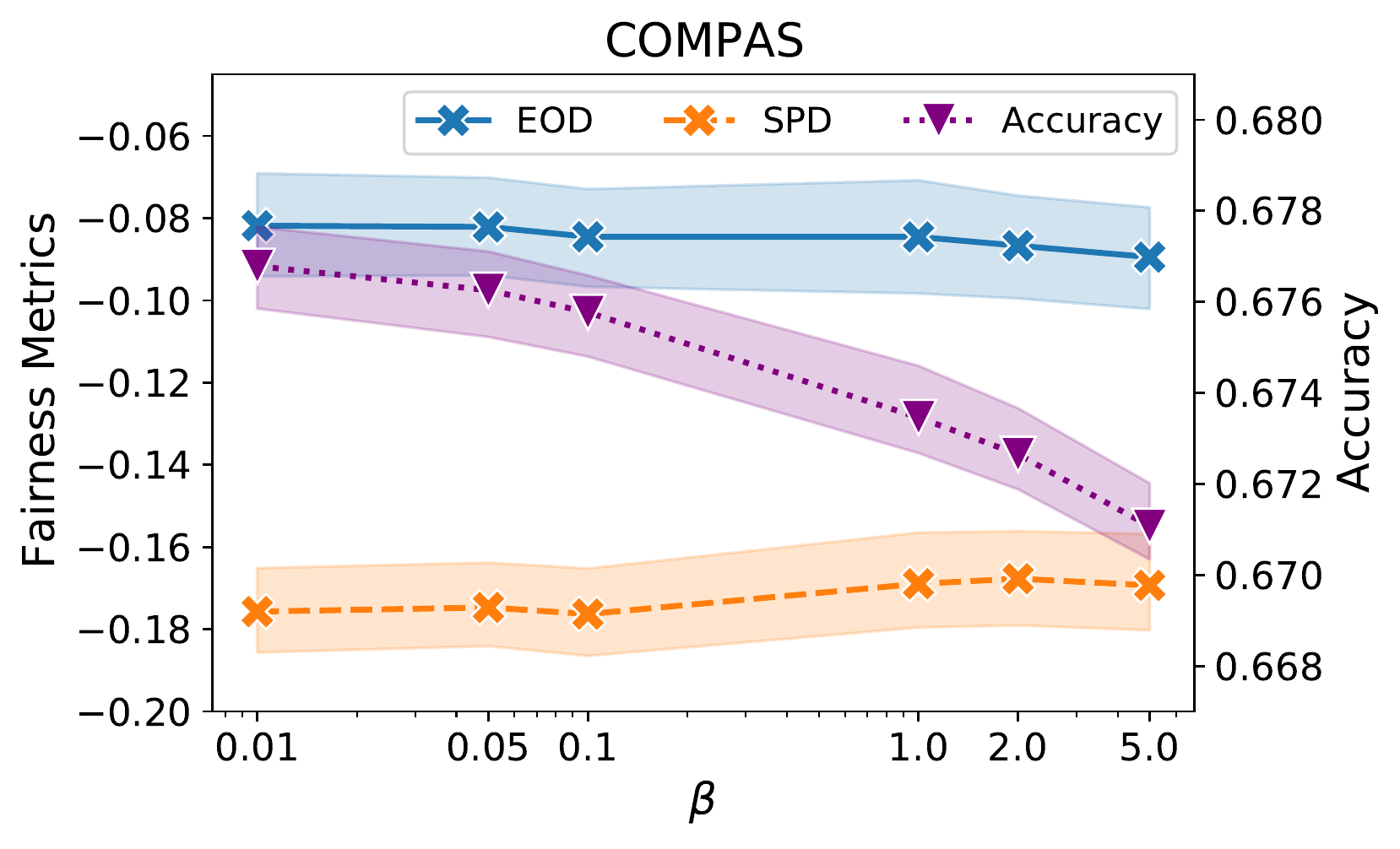}}
    \vspace{-1.5em}
    \caption{Effects of fairness budget $\beta$ for 5 clients and heterogeneity $\alpha=0.2$ on FairFed with local reweighting.}
    \label{fig:beta}
\end{figure}

\subsubsection{Performance with different fairness budgets ($\beta$).}
In FairFed, we introduced a fairness budget parameter $\beta$, which controls how much the aggregation weights can change due to fairness adaptation at the server in each round (refer to~\eqref{eq:FairAgg_weight_def} for the explanation of $\beta$). Figure \ref{fig:beta} visualizes the effects of $\beta$ using heterogeneity level $\alpha = 0.2$. As the value of $\beta$ increases, the fairness constraint has a bigger impact on the aggregation weights, yielding better fairness (EOD closer to zero)  at the cost of a decrease in model accuracy.

\section{Cases Studies for Fair Training in FL}\label{sec:case_studies}
In Section~\ref{sec:experimental_evaluation}, we evaluated FairFed on heterogeneous distributions synthesized from standard benchmark datasets in fair ML. In order to validate the effectiveness of FairFed in fair FL scenarios with naturally heterogeneous distributions, we consider two FL case studies in this section.
\subsection{Case Study 1: Predicting individual income from US Census across states}

In this case study, we use the US Census data to present the performance of our FairFed approach (introduced in Section~\ref{sec:fairfed}) in a distributed learning application with a natural data partitioning. 
Our experiments are performed on the ACSIncome dataset~\cite{ding2021retiring} with the task of predicting whether an individual's income is above \$50,000 (or not) based on the features collected during the Census which include employment type, education, martial status, etc. 
\begin{figure}[h]
    \subfigure{
    \includegraphics[width=0.45\columnwidth]{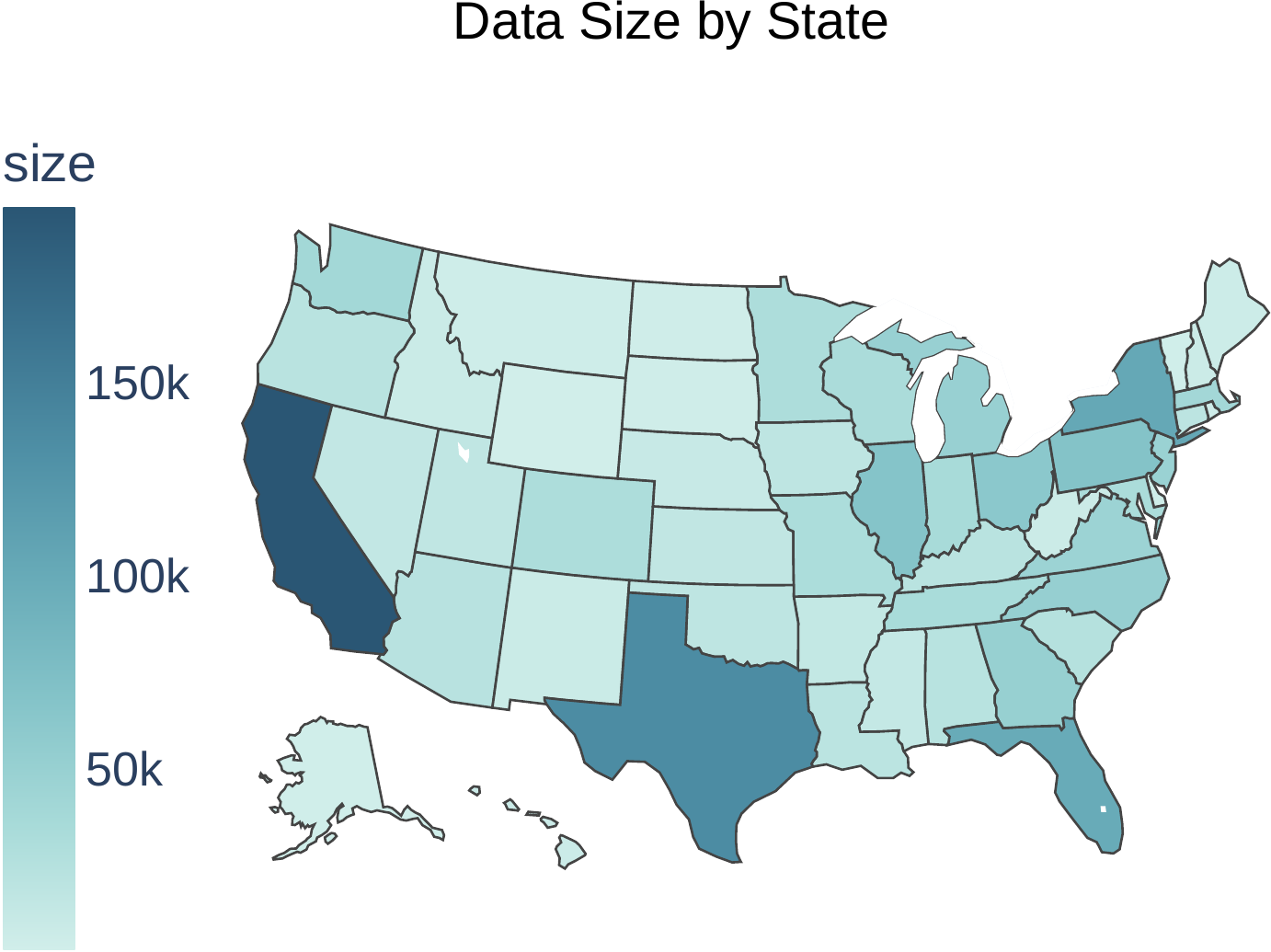}}
    \subfigure{
    \includegraphics[width=0.45\columnwidth]{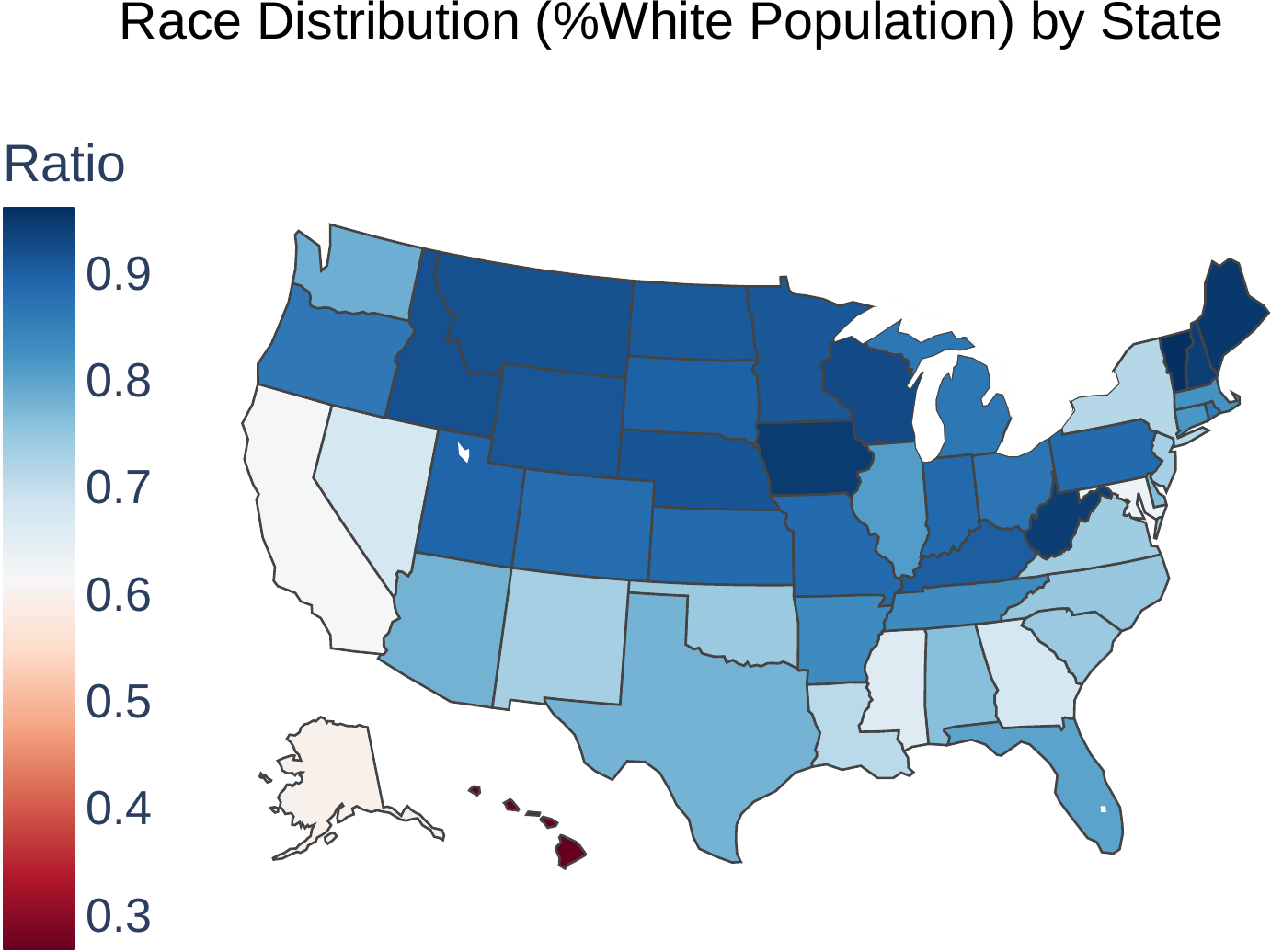}}
    \caption{Demographic distribution of ACSIncome dataset.}
\label{fig:acs_distribution}
\end{figure}

\subsubsection{Data Distribution.} ACSIncome dataset is constructed from American Community Survey (ACS) Public Use Microdata Sample (PUMS) over all 50 states and Puerto Rico in 2018 with a total of 1,664,500 datapoints. In our experiments, we treat each state as one participant in the FL system (i.e., 51 participants). Due to the demographic distribution of different states, clients share different data sizes and sensitive attributes distribution. For example, Wyoming has the smallest dataset size with 3,064 users compared to California that has 195,665 users. We choose the race information (white/non-white) of the users as the sensitive attribute of interest in our experiments. Hawaii is the state with the lowest ratio (26\%) of white population. , while Vermont has the highest ratio (96\%) of its dataset as white population. Figure~\ref{fig:acs_distribution} provides a visualization for the data distributions across the different states. 

\subsubsection{Performance of FairFed.} Table \ref{tab:acsincome} compares the performance of FairFed on ACSIncome dataset. Table~\ref{tab:acsincome} shows that adopting local reweighting yields worse group fairness performance than simply applying FedAvg (without any debiasing) due to the heterogeneity across states. FairFed with reweighting overcomes this issue and improves the EOD by 20\% (-0.062 to -0.050). 


\subsection{Case Study 2: predicting daily stress level from wearable sensor signals}
In this case study, we use the human behavior dataset TILES~\cite{mundnich2020tiles}. \textit{Tracking Individual Performance with Sensors} (TILES) is a 10 weeks longitudinal study with hospital worker volunteers in a large Los Angeles hospital, where 30\% of the participants are male and 70\% are female. We use this dataset to estimate users' daily stress levels based on physiological and physical activity signals collected through wearable sensors (e.g., Fitbit) . The target is a binary label indicating whether the person's stress level is above individual average (i.e., 1) or not (i.e., 0), which are collected from daily surveys sent to the participants' phones. 

\smallskip

\subsubsection{\bf Data Distribution.}
We focus on the nurse population in the dataset. Each client represents the data from one occupation group -- day-shift registered nurse (RN- day shift), night-shift registered nurse (RN-night shift), and certified nursing assistant (CNA). The three clients vary by data size, the distributions of gender (the sensitive attribute) and target stress variable. In general, the client of day-shift registered nurse population has the most datapoints, more female, and higher stress levels. The detailed data distribution of each clients is shown in Table \ref{tab:tiles_distribution}. 
\begin{table}[]\small
    \centering
    \begin{tabular}{c|c||c|c||c|c}
    \hline
    \multirow{2}{*}{\bf Client} &  \multirow{2}{*}{\bf Size} & \multicolumn{2}{c||}{\bf Gender} & \multicolumn{2}{c}{
    \bf Stress} \\ 
    & & F & M & $y=0$ & $y=1$ \\\hline
     RN-day shift & 707 & 82\% & 18\%  & 45\% & 55\%\\\hline
     RN-night shift & 609 & 77\% & 23\% & 57\% & 43\%\\\hline
     CNA & 244 & 61\% & 39\% & 65\% & 35\% \\ 
    \hline
    \hline
    \end{tabular}
    \caption{Data distribution of TILES dataset.}
    \label{tab:tiles_distribution}
\end{table}

\subsubsection{Performance of FairFed.} Table \ref{tab:acsincome} reports the performance on TILES dataset. Both FairFed and local reweighting improve the EOD metric as compared to FedAvg. FairFed improves EOD from -0.199 to 0.004 with only 2.6\% accuracy decrease (from 0.567 to 0.556). 

\begin{table}\small
\centering
\setlength{\tabcolsep}{2pt}
\begin{tabular}{c|c c c | c c c}
    \hline
    \hline
      \multirow{2}{*}{Method}& \multicolumn{3}{c|}{ACSIncome} & \multicolumn{3}{c}{TILES} \\
      & Acc. & EOD & SPD & Acc. & EOD & SPD \\
      \hline
         FedAvg & 
         \textbf{0.800} & -0.062	& -0.102 & \textbf{0.567}	& -0.199	&-0.166 \\
         Local / RW &  0.800	& -0.066 & -0.106 & 0.567	& -0.064&-0.041\\
         FairFed / RW  & 0.799 &	\textbf{-0.050} &	\textbf{-0.089} & 0.556 &	\textbf{0.004} &\textbf{	0.004}\\
    \hline
    \hline
    \end{tabular}
    \caption{Performance on ACSIncome and TILES datasets.}
    \label{tab:acsincome}
\end{table}




\section{Conclusion and Future Works}
In this work, motivated by the importance and challenges of group fairness in federated learning, we propose the FairFed algorithm to enhance group fairness via a fairness-aware aggregation method, aiming to provide fair model performance across different sensitive groups (e.g., racial, gender groups) while maintaining high utility. Though our proposed method outperforms the state-of-the-art fair federated learning frameworks under high data heterogeneity, limitations still exist. As such, we plan to further improve FairFed from these perspectives: 1) We report the empirical results on binary classification tasks in this work. We will extend the work to various application scenarios (e.g., regression tasks, NLP tasks); 2) We will extend our study to scenarios of heterogeneous application of different local debiasing methods and understand how the framework can be tuned to incorporate updates from these different debiasing schemes; 
3) We currently focus on group fairness in FL, but we plan to integrate FairFed with other fairness notions in FL, such as \textit{collaborative fairness} and \textit{client-based fairness}.  
We give an example of how FairFed can incorporate both group-fairness and client-based fairness in Appendix~\ref{sec:modified_fairfed}, which sets promising preliminary steps for future exploration.

\bibliography{ref.bib}

\appendix

\clearpage

\section{Changes to FairFed when tracking the Statistical Parity Difference metric}\label{app:SPD}
In order for the server to compute the global Statistical Parity Difference (SPD) metric from model evaluation information shared by the clients, we follow a decomposition of the SPD metric similar to what we did for EOD in Section~\ref{sec:computing_global_metric_EOD}. In particular, we can decompose the SPD metric defined in~\eqref{eq:SPD} as:
\begin{align}
\label{eq:global_component_SPD}
&F_{global} = SPD \nonumber = \Pr(\hat{Y}\!=\!1|A\!=\!0) \!-\! \Pr(\hat{Y}\!=\!1|A\!=\!1)\nonumber\\
 &\!\!\stackrel{(a)}=\!\!\sum_{k=1}^{K}\!\frac{n_k}{n} \Bigg[ \tfrac{\Pr(\hat{Y}=1|A=0,C=k)\Pr(\!A=0|C=k)}{\Pr(A=0)} \nonumber \\
 &\hspace{2.5em}\underbrace{\hspace{1.5em} \qquad - \tfrac{\Pr(\hat{Y}=1|A=1,C=k)\Pr(\!Y=1|C=k)}{\Pr(A=1)}\Bigg]}_{m_{global,k}^{(SPD)}}\!, \nonumber \\
 &=\sum_{k=1}^{K} m_{global,k}^{(SPD)},
\end{align}
where in $(a)$ we substitute the probability of a datapoint being sampled at client $k$ with the ratio of points at client $k$ to the total number of points.

Thus, from~\eqref{eq:global_component_SPD}, when FairFed is tracking SPD, the FairFed framework summarized in Algorithm~\ref{alg:fairagg}is in the following two places: (1) The server needs to securely aggregate the statistics $\mathcal{S} = \left\{\Pr(A=0), \Pr(A=1)\right\}$ in the \textit{Dataset statistics step} of Algorithm~\ref{alg:fairagg}; (2) Each client will compute $m_{global,k}^{(SPD)}$ using the dataset statistics $\mathcal{S}$ and their local evaluation of the global model as defined in~\eqref{eq:global_component_SPD} above.
~
\section{Heterogeneity of sensitive attribute distribution in our experiments}
\label{app:datasets}
In this work, we use a generic non-IID synthesis method based on the Dirichlet distribution proposed in~\citep{sampling_non_iid_google} to control the heterogeneity of the sensitive attribute distribution at each client. Table \ref{tab:hetero_example} shows an example of the heterogeneous data distribution for the Adult and COMPAS datasets for $\alpha=0.1$ and $10$.


\begin{table}[ht] \small
\centering
\renewcommand{\arraystretch}{1.1}
    \begin{tabular}{c|c|c||c|c}
    \multicolumn{5}{c}{\bf Adult Dataset}\\
    \hline
    \multirow{2}{*}{\bf Client ID} & \multicolumn{2}{c}{$\alpha=0.1$} & \multicolumn{2}{c}{$\alpha=10$} \\ 
    & $A=0$ & $A=1$ & $A=0$ & $A=1$ \\\hline
     0 & 269 & 615 & 1505 & 3585 \\\hline
     1 & 128 & 29839 & 876 & 5695 \\\hline
     2 & 418 & 74 & 978 & 7261 \\\hline
     3 & 43 & 392 & 601 & 5848 \\\hline
     4 & 4196 & 203 & 1094 & 8734 \\\hline
    \hline
    \end{tabular}\hspace{6mm}
    \begin{tabular}{c|c|c||c|c}
    \multicolumn{5}{c}{\bf COMPAS Dataset}\\
    \hline
    \multirow{2}{*}{\bf Client ID} & \multicolumn{2}{c}{$\alpha=0.1$} & \multicolumn{2}{c}{$\alpha=10$} \\ 
    & $A=0$ & $A=1$ & $A=0$ & $A=1$ \\\hline
     0 & 32 & 423 & 612 & 217 \\\hline
     1 & 151 & 411 & 876 & 109 \\\hline
     2 & 3 & 62 & 836 & 185 \\\hline
     3 & 522 & 42 & 880 & 251 \\\hline
     4 & 3286 & 1 & 790 & 177 \\\hline
    \hline
    \end{tabular}
    \caption{\label{tab:hetero_example}An example of the heterogeneous data distribution (non-IID) on the sensitive attribute $A$ (sex) used in experiments on the Adult and COMPAS datasets for $K=5$ and heterogeneity parameters $\alpha=0.1$ and $\alpha=10$.}
\end{table}

\section{Additional experimental results}
\label{app:more_results}

\subsubsection{Performance under different local debiasing strategies.}
One of the notable advantages of FairFed is that each client can have the freedom of adopting different local debiasing strategies. For instance, a fraction of clients might not adopt any local debiasing strategy due to inavailability of the full local dataset at any instance (online stream of datapoints). In this section, we highlight the performance of FairFed under such realistic application scenario. 
\begin{figure}[t!]
\centering
    \includegraphics[width=0.95\columnwidth]{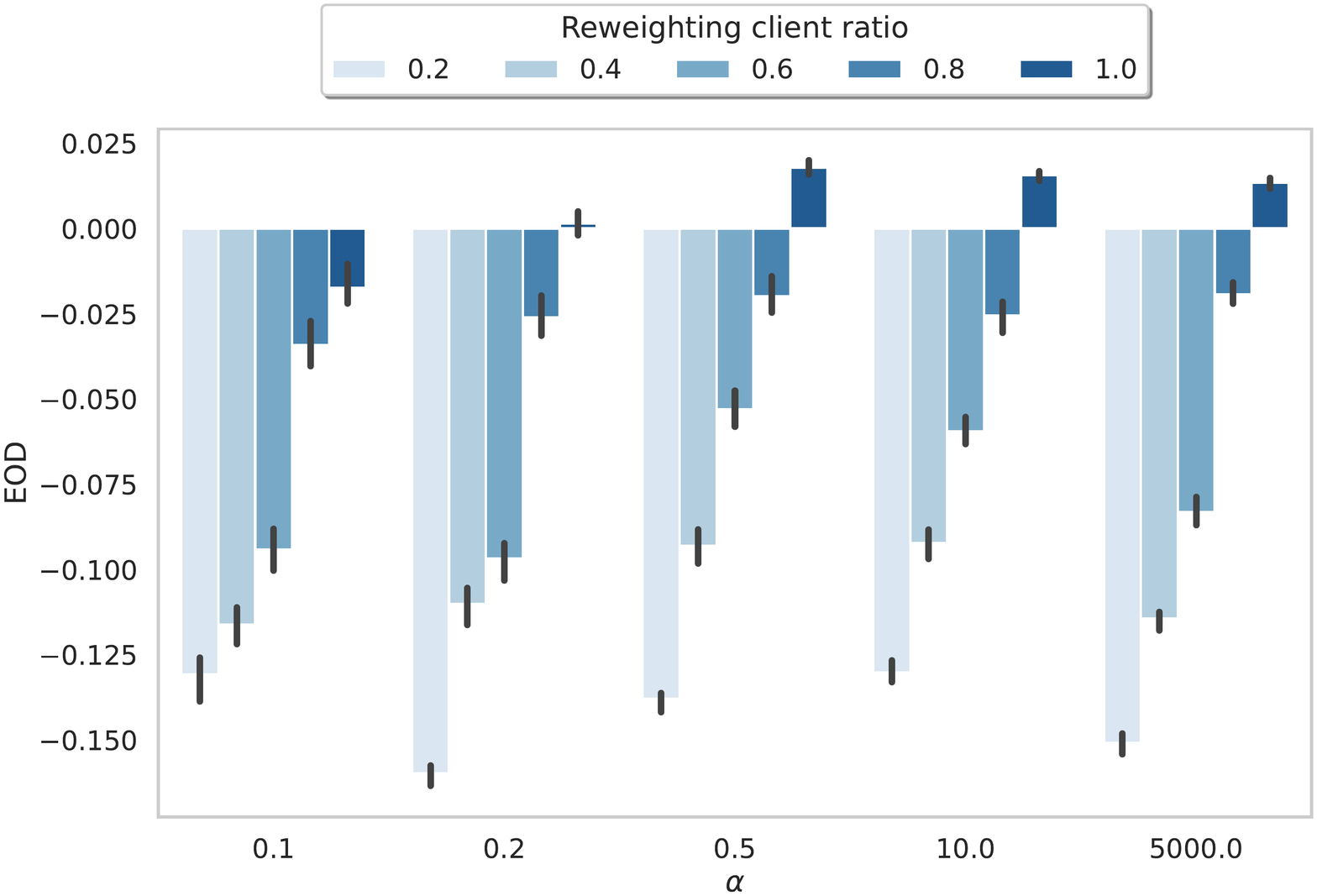}
   \caption{Effect of only a subset of clients adopting the reweighting debiasing method.}
    \label{fig:ratio_effect}
\end{figure}
We empirically study this for Adult dataset where we simulate a scenario in which only a fraction of clients adopt the local reweighting strategy under FairFed. The remaining clients will participate in the FL system without local debiasing, but they still have to communicate their local fairness metrics and global fairness components $m_{global,k}$ as described in Algorithm~\ref{alg:fairagg}. 

As seen in Figure~\ref{fig:ratio_effect}, the group fairness improves as more clients adopt the local debiasing strategy. This also highlights the importance of both components of FairFed as the fairness-aware aggregation is not individually sufficient to achieve group fairness. In particular, FairFed needs 40\% of the clients to use local debiasing in order to outperform local reweighting without fairness-aware aggregation. When more than 60\% clients adopt local reweighting, FairFed outperforms both FedAvg with local reweighting and global reweighting methods. The minimum accuracy across the different ratios and $\alpha$ was 0.826 suggesting a small accuracy penalty.

\begin{figure}[h!]
\centering
    \includegraphics[width=0.95\columnwidth]{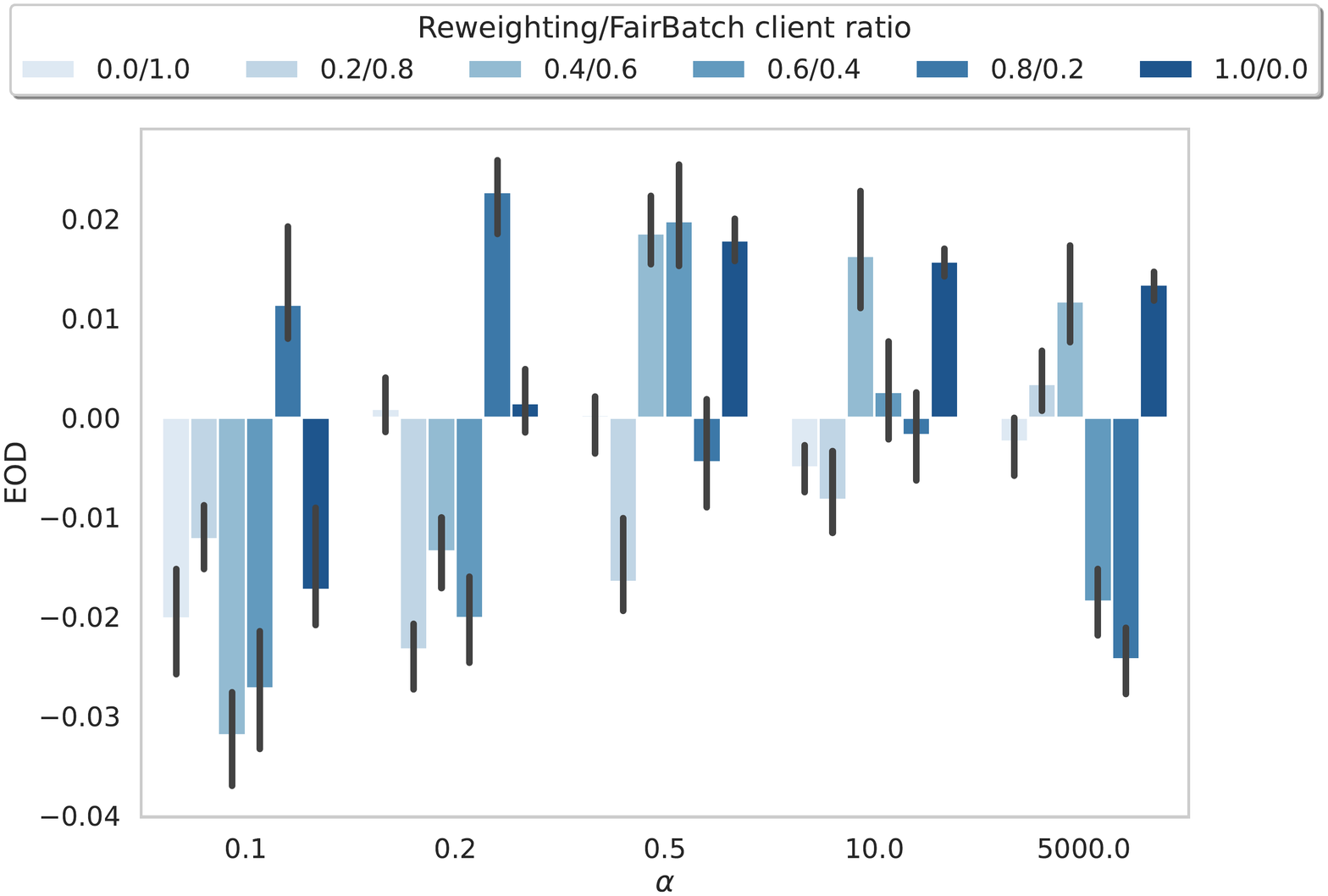}
   \caption{Effects of different local debiasing strategies (mixture of reweighting and FairBatch clients).}
    \label{fig:mixed_effect}
\end{figure}

We also conducted experiments to verify the innate flexibility of FairFed to operate with a mixed application of debiasing techniques. In particular, we conduct experiments where a subset of clients apply the reweighting debiasing method, while the remaining clients apply the FairBatch debiasing method~\citep{roh2021fairbatch}. 
From Figure~\ref{fig:mixed_effect}, we see that even when different debiasing methods across clients are employed, FairFed still achieves an improvement in the fairness metrics as compared to local/no debiasing baselines (details for each baseline shown in  Table~\ref{tab:compare_partition}). The minimum accuracy across the different ratio, was 0.829, which is an accuracy loss of at most 1.1\%.

This also compliments our observation for Figure~\ref{fig:ratio_effect}, which shows a huge degradation in the fairness metric as the fraction of FairFed clients applying local reweighting decreased. Our results in Figure~\ref{fig:mixed_effect} emphasizes that this great reduction can be greatly reduced by applying other debiasing algorithm at these clients and not necessarily reweighting as the local debiasing algorithm of choice.

\subsubsection{Performance on clients with a single sensitive group.}
In order to verify how FairFed works in scenarios where a client's local fairness metric ($F_k$ in~\eqref{eq:FairAgg_weight_def}) cannot be computed, we construct an experiment on the Adult dataset where a client is comprised completely from a single group. In particular, we consider 5 clients, where the first two are comprised of only female datapoints and remaining three contain only male points. We use the aforementioned  Dirichlet distribution (with $\alpha=0.5$) to partition each group, based on the target prediction variable (Income $>$ 50k), into their corresponding subset of clients (first two containing all female datapoints, and the remaining three containing only males). An example of such data partitioning is presented in Table~\ref{tab:single_group_clients}. 

\begin{figure}[t!]
\centering
    \includegraphics[width=0.95\columnwidth]{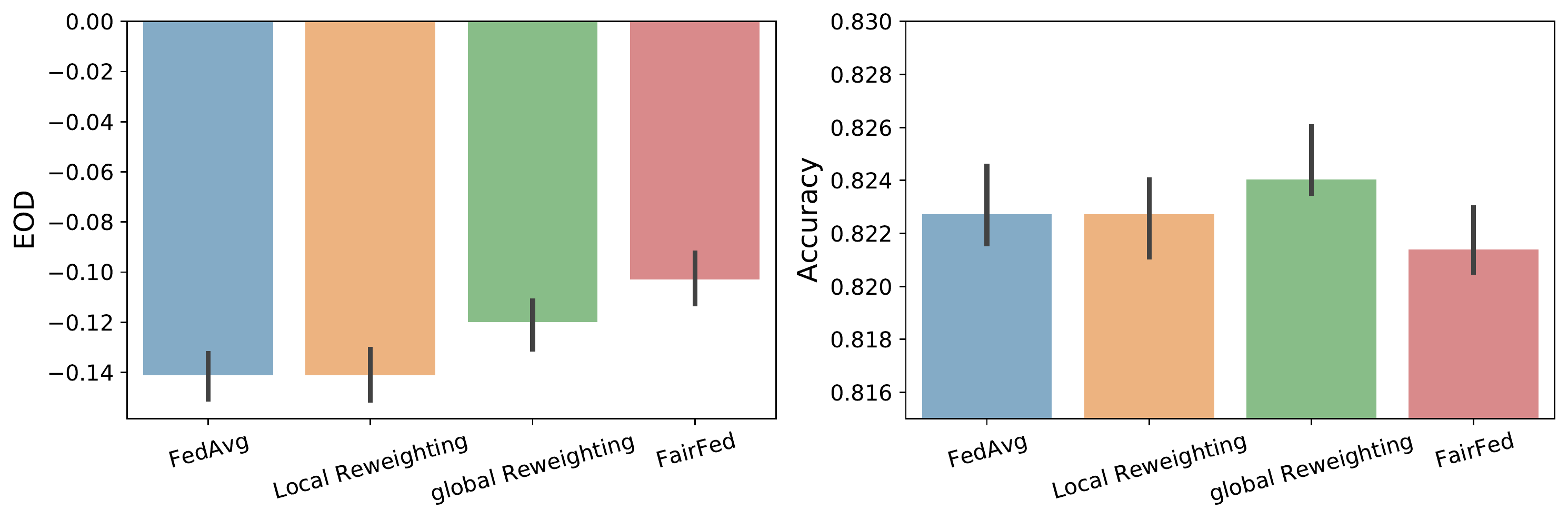}
   \caption{Performance with clients that only contain data from one sensitive group.}
    \label{fig:single_group}
\end{figure}

Figure~\ref{fig:single_group} compares the performance of FairFed using reweighting as with the reweighting baselines (local and global). Since each client contains only a single group, then local reweighting is ineffective and performs similar to the FedAvg baseline. Both FairFed and global reweighting improve the EOD metric as they take into account the datapoints across the different clients. In particular, FairFed improves over FedAvg by 27\%, while global reweighting improves EOD by 15\%.

\begin{table}[t] \small
\centering
    \begin{tabular}{c|c||c|c}
    \multicolumn{4}{c}{\bf Adult Dataset (Single Group per Client Experiment)}\\
    \hline
    \multirow{2}{*}{\bf Client ID} &
    \multirow{2}{*}{\bf Gender} &
    \multicolumn{2}{c}{\bf Target: Income $>$ 50k} \\ 
    & & $y=0$ & $y=1$  \\\hline
     0 & (100\%)Female & 3722 & 662 \\\hline
     1 & (100\%)Female & 3917 & 299 \\\hline
     2 & (100\%)Male & 1641 & 2888 \\\hline
     3 & (100\%)Male & 5815 & 2409 \\\hline
     4 & (100\%)Male & 4660 & 36 \\\hline
    \hline
    \end{tabular}\vspace{2mm}
    \caption{\label{tab:single_group_clients} An example of the heterogeneous data distribution ($\alpha=0.5$) for $K=5$ clients on the target variable  used in the experiment on the Adult dataset, where each client is assigned only points with a single sensitive attribute value. }
\end{table}

\section{FairFed with Uniform Accuracy constraints}\label{sec:modified_fairfed}
Beyond optimizing group fairness metrics, the notion of client-based fairness is critically important in trustworthy FL systems, since FL clients would be more reluctant to collaborate if they end up with models that perform badly on their local data distributions.
To pursue the goal of client-based fairness in parallel to group fairness, our proposed FairFed algorithm can also be revised to incentivize uniform accuracy across clients. \textit{Uniform accuracy} aims to minimize the performance differences across client. To quantify this criteria, we calculate the standard deviation of the accuracy (Std-Acc.) of each client as the metric of \textit{uniform accuracy}, where smaller values indicates better uniformity.
The revised procedure (i.e., revised version of \eqref{eq:FairAgg_weight_def}) computes the aggregation weights as follows for $\forall k \in [K]$:
\begin{align}\label{eq:FairAgg_revise}
    \Delta_k^t &=
    \begin{cases}
      \left|Acc_k^t - \overline{Acc^t}\right| & \text{if $F_k^t$ is undefined} \\\eta|F^t_{\rm global} - F_k^t| & \\ + (1-\eta)|Acc_k^t - \overline{Acc^t}| & \text{otherwise}
    \end{cases}, 
     \nonumber\\
    \bar{\omega}_k^t &= \bar{\omega}_k^{t-1} {-} \beta \!\!\left(\!\!\Delta_k - \frac{1}{K}\!\sum_{i=1}^K \Delta_i\!\!\right)\!,\quad  {\omega}_k^t = \frac{\bar{\omega}_k^t}{ \textstyle\sum_{i=1}^K \bar{\omega}_i^t}. 
\end{align}
We introduce a parameter $\eta\in[0,1]$, which controls the trade-off between global fairness constraint and uniform accuracy constraint on the weight update. When $\eta$ is large, the global fairness constraint has higher impact on aggregation weights update. In particular, when $\eta = 1$, we recover the FairFed functionality presented in~\eqref{eq:FairAgg_weight_def}.
\begin{table*} \footnotesize
\centering
    \begin{tabular}{c|c |a c c c c | a c c c c}
    \hline
    \hline
      & \multirow{3}{*}{Method}& \multicolumn{5}{c|}{Adult} & \multicolumn{5}{c}{COMPAS}  \\
      \cline{3-12}
          &    & \multicolumn{5}{c|}{Heterogeneity Level $\alpha$} & \multicolumn{5}{c}{Heterogeneity Level $\alpha$}\\
         & & 0.1 & 0.2 & 0.5 & 10 & 5000 & 0.1 & 0.2 & 0.5 & 10 & 5000 \\
         \hline
          \multirow{4}{*}{Acc.} 
            & FedAvg &  \textbf{0.835} &  \textbf{0.836} &  \textbf{0.835} &  \textbf{0.836} &  \textbf{0.837} &  \textbf{0.674} &  \textbf{0.673} &  \textbf{0.675} &  \textbf{0.674} &  \textbf{0.675} \\
            & Local / RW &  0.834 &  0.834 &  0.832 &  0.829 &  0.829 &  0.674 &  0.673 &  0.675 &  0.676 &  0.675 \\ 
          & FairFed / RW ($\eta = 0.2$) &0.829&0.829&0.828&0.826&0.825 & 0.669&0.681&0.677&0.671&0.669\\
            & FairFed / RW ($\eta = 1.0$) &  0.830 &  0.834 &  0.832 &  0.829 &  0.829 &  0.672 &  0.670 &  0.669 &  0.669 &  0.673 \\
        \hline
        \multirow{4}{*}{EOD} 
        & FedAvg & -0.174 & -0.173 & -0.176 & -0.179 & -0.180 & -0.065 & -0.071 & -0.067 & -0.076 & -0.078 \\
        & Local / RW & -0.065 & -0.042 &  0.017 &  0.015 &  0.014 & -0.062 & -0.069 & -0.067 & -0.074 & -0.077 \\ 
          & FairFed / RW ($\eta = 0.2$) & -0.058&-0.005&0.020&0.021&0.021&-0.085&-0.085&-0.085&-0.088&-0.082
          \\
        & FairFed / RW ($\eta = 1.0 $)&   \textbf{-0.017} &  \textbf{0.001} &  \textbf{0.018} &  \textbf{0.016} &  \textbf{0.013} & \textbf{-0.057} & \textbf{-0.065} & \textbf{-0.053} & \textbf{-0.067} & \textbf{-0.061} \\           
          \hline
            
    \multirow{5}{*}{Std-Acc.} & FedAvg & 0.082&0.070&0.067&0.021&0.01& 0.053&0.060&0.046&0.034&0.0\\
          & Local / RW &0.085&0.074&0.070&0.022&0.010&
          0.054&0.060&0.045&0.034&0.0\\
          & FairFed / RW ($\eta = 0.2$) & \textbf{0.062}&\textbf{0.060}&\textbf{0.052}&\textbf{0.020}&\textbf{0.009}&
          \textbf{0.049}&\textbf{0.048}&\textbf{0.031}&\textbf{0.032}&0.0
          \\
          &  FairFed / RW ($\eta = 1.0 $)&  0.064&0.062&0.055&0.020&0.009&  
          0.054&0.055&0.043&0.032&0.0\\
          
    \hline
    \hline
    \end{tabular}
    \vspace{2mm}
    \caption{Performance comparison of uniform accuracy constraint $\eta$ on data partition with different heterogeneity levels $\alpha$. We report the average performance of 20 random seeds.
    }
    \label{tab:compare_partition_eta}
\end{table*}
\subsection{Experimental results with synthetic partitioning}
We evaluate the performance of the revised FairFed using experiments with the same configurations as Section \ref{sec:config}. Table \ref{tab:compare_partition_eta} compares the performance with different data heterogeneity and reweighting-based debiasing methods. For both Adult and COMPAS datasets, FairFed improves EOD and Std-Accuracy metrics simultaneously, while local debiasing can exaggerate the accuracy differences across clients. In particular, in highly heterogeneous scenarios distributions (i.e., with smaller $\alpha$), the performance variances across clients are large; It can be seen that in these settings, FairFed provides better improvement to the uniform accuracy metric.

\subsubsection{\bf Effects of parameter $\eta$.}
\begin{figure}[!t]
\centering
    \subfigure{
    \includegraphics[width=0.47\columnwidth]{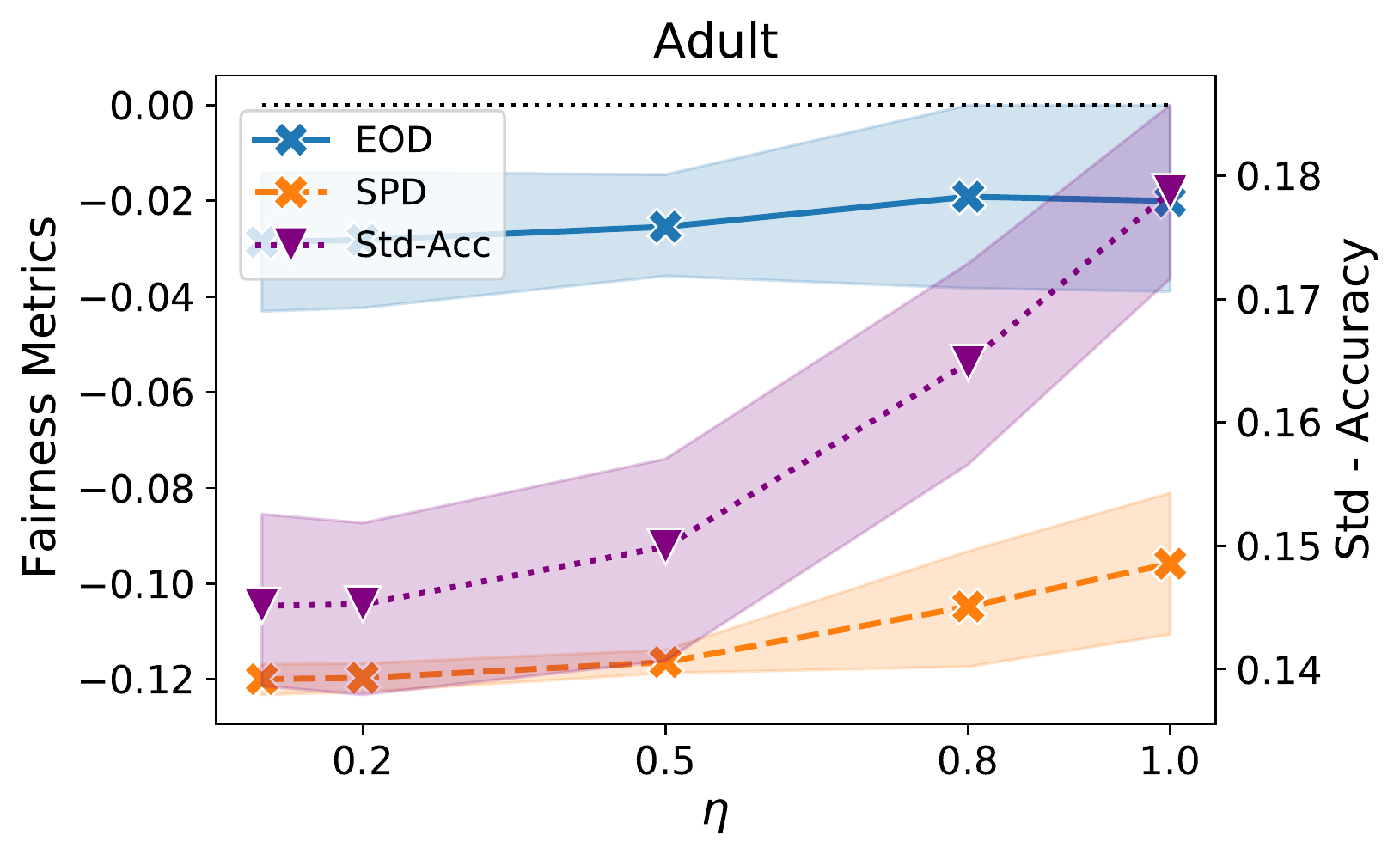}}
    \subfigure{
    \includegraphics[width=0.47\columnwidth]{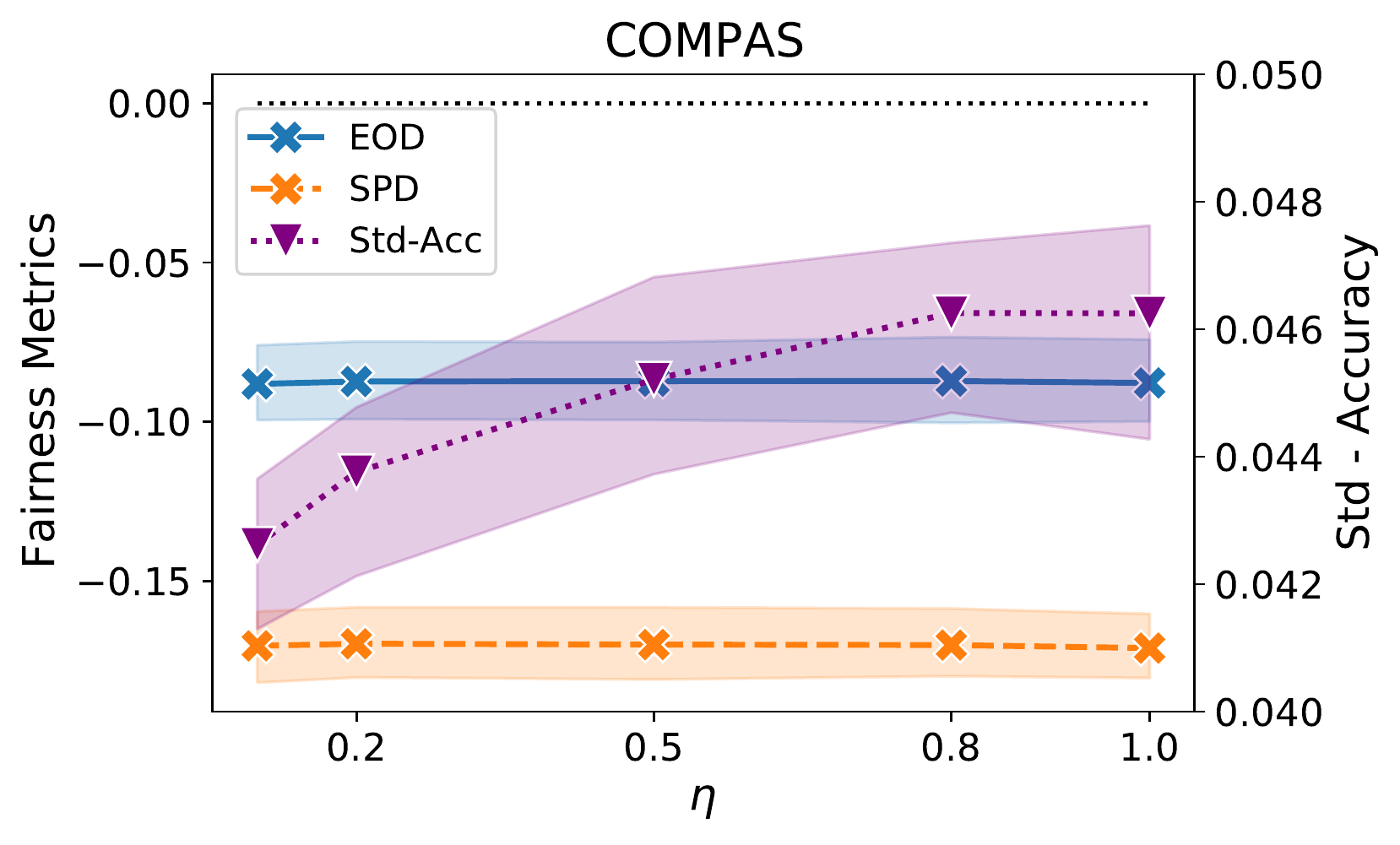}}
    \caption{Effects of parameter $\eta$ on the performance of FairFed for $K=5$ clients and heterogeneity parameter $\alpha=0.5$.}
    \label{fig:eta}
\end{figure}

Figure \ref{fig:eta} illustrates the effect of $\eta$ using heterogeneity level $\alpha = 0.5$ as an example. We use \textit{Std-Acc.} as a measurement for the uniformity of accuracy. As $\eta$ increases, the fairness constraint has a bigger impact on the aggregation weights, yielding a better group fairness performance as a trade-off for accuracy uniformity.
\begin{figure}[t!]
\centering
\subfigure{\label{fig:acsincome_eta}
\includegraphics[width=0.47\columnwidth]{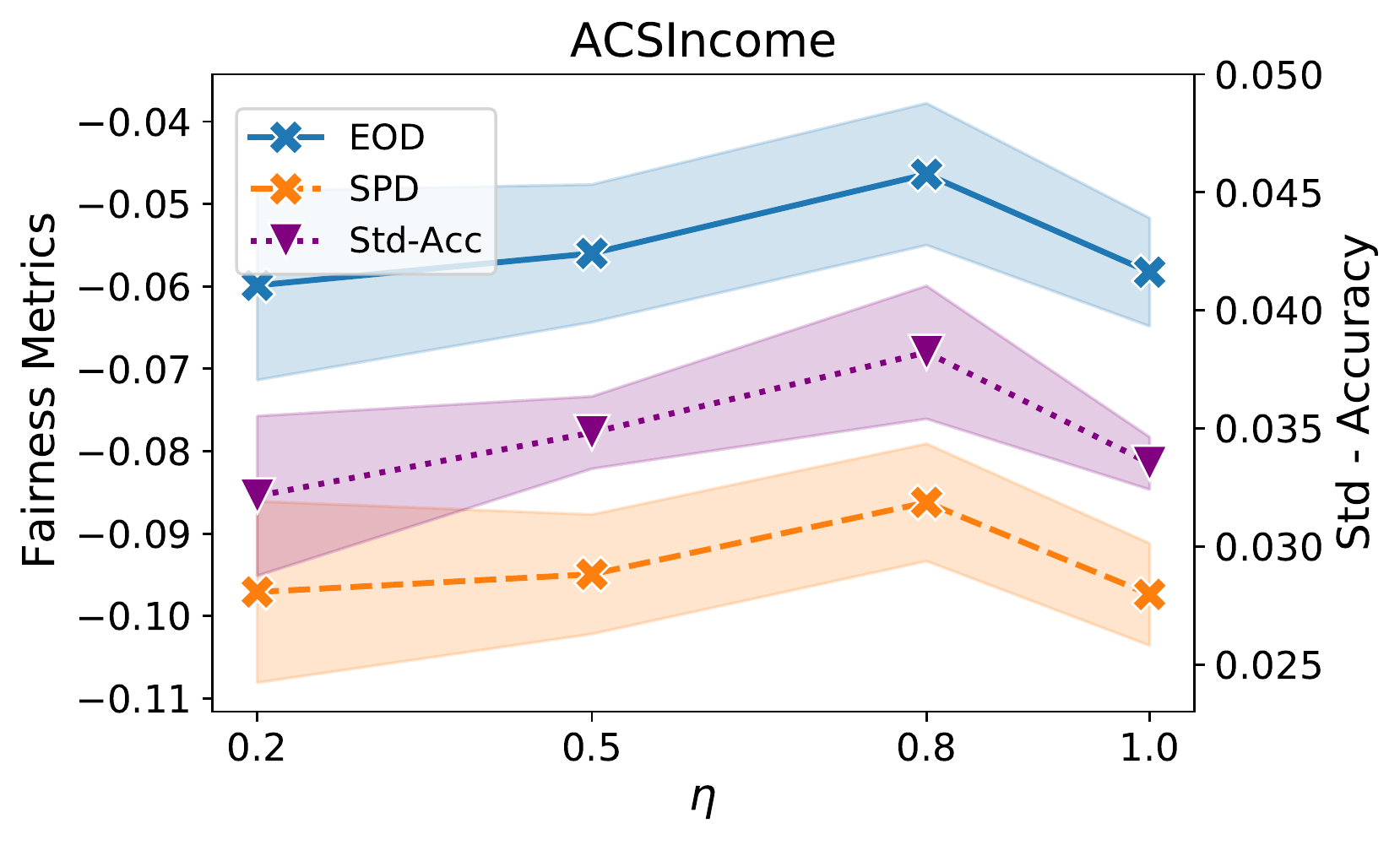}
}
\subfigure{\label{fig:tiles_eta}
\includegraphics[width=0.47\columnwidth]{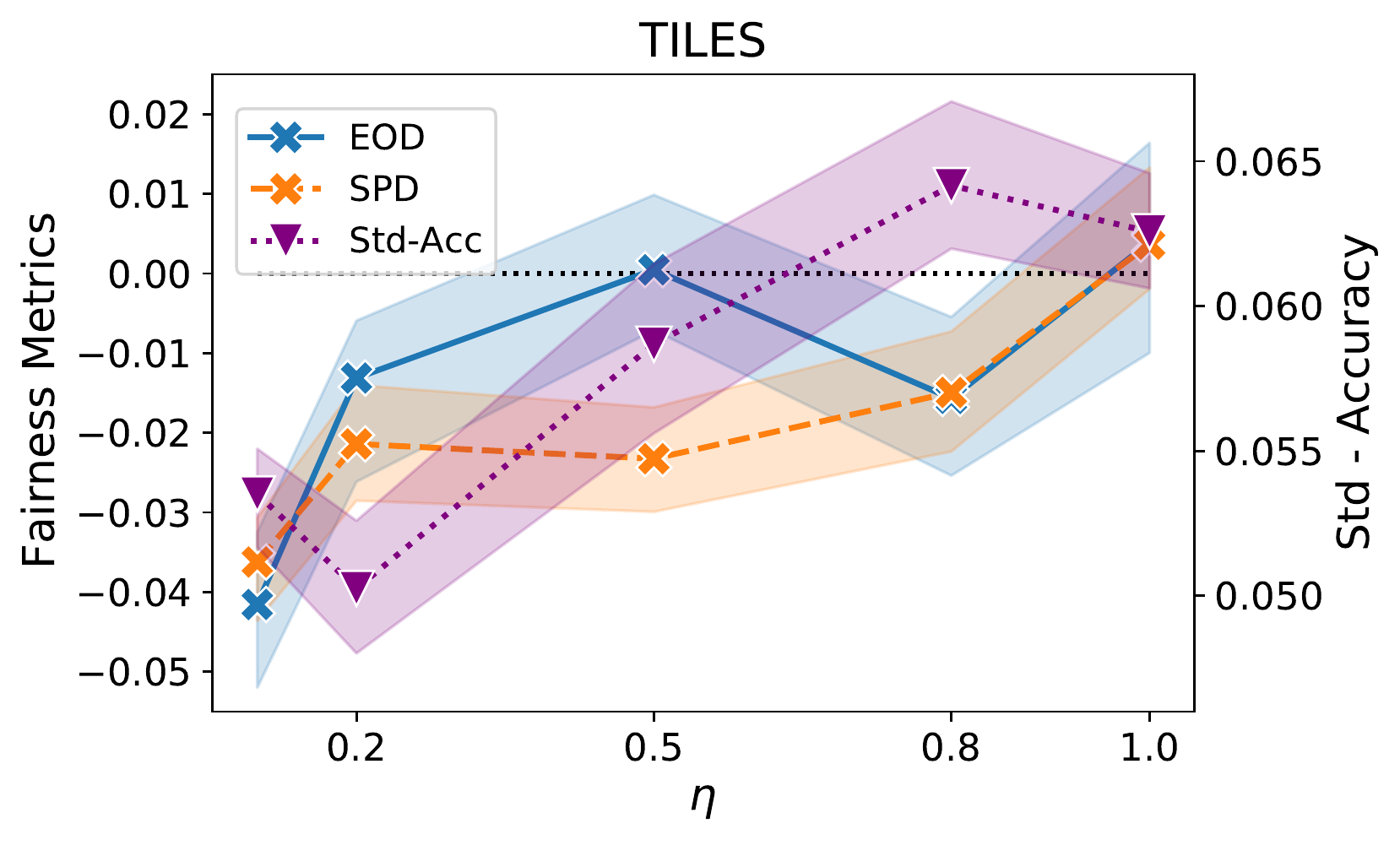}
}
\caption{Effects of $\eta$ on the ACS Income and TILES datasets.}
\label{fig:acs_tiles_eta_fig}
\end{figure}
\subsection{Experimental results FL case studies}
We also evaluated the performance of FairFed with uniform accuracy constraints on the ACSIncome and TILES datasets studied in Section~\ref{sec:case_studies}.

In these datasets, the uniformity of accuracy of different clients can also be controlled by uniform accuracy constraint (i.e., $\eta$); Figure \ref{fig:acs_tiles_eta_fig} visualizes the trend that when $\eta$ decreases, the accuracy uniformity across clients improved with the trade-off of group fairness metrics.


\end{document}